\documentclass[12pt]{article}
\usepackage{macros}

\begin{document}

\title{Adaptive Querying with AI Persona Priors}\blfootnote{Authors are listed in alphabetical order. Code and data are publicly available at \url{https://github.com/yw3453/adaptive-query-ai-persona-priors}.
}

\author{
    Kaizheng Wang\thanks{Department of Industrial Engineering and Operations Research and Data Science Institute, Columbia University.} \quad
    Yuhang Wu\thanks{Decision, Risk, and Operations Division, Columbia Business School.} \quad
    Assaf Zeevi\footnotemark[2]
}

\date{This version: May 30, 2026}

\maketitle
\vspace{-1em}

\begin{abstract}
We study adaptive querying for learning user-dependent quantities of interest, such as responses to held-out items and psychometric indicators, within tight query budgets. Classical Bayesian design and computerized adaptive testing typically rely on restrictive parametric assumptions or expensive posterior approximations, limiting their use in heterogeneous, high-dimensional, and cold-start settings. We introduce a persona-induced latent variable model that represents a user's state through membership in a finite dictionary of AI personas, each offering response distributions produced by a large language model. This yields expressive priors with closed-form posterior updates and efficient finite-mixture predictions, enabling scalable Bayesian design for sequential item selection. Experiments on synthetic data and WorldValuesBench demonstrate that persona-based posteriors deliver accurate probabilistic predictions and an interpretable adaptive elicitation pipeline.
\end{abstract}


\section{Introduction}

Many interactive systems must learn about users under severe information constraints. Examples range from market research surveys and psychometrics to recommender systems and preference elicitation, where only a small number of questions can be asked before user fatigue, privacy concerns, or cost constraints intervene. In these settings, the goal is not merely to predict individual responses, but to form calibrated probabilistic beliefs about user-dependent quantities of interest---such as held-out responses, psychometric indicators, or downstream decisions---within tight query budgets. Bayesian adaptive querying provides a natural framework for this problem by explicitly modeling uncertainty and selecting questions to maximally reduce it.

Despite its conceptual appeal, existing Bayesian design and adaptive testing methods face a practical tension between \emph{expressiveness} and \emph{tractability}. Classical computerized adaptive testing (CAT) and item response theory (IRT) typically rely on low-dimensional parametric latent traits, which can be restrictive when response patterns are heterogeneous and high-dimensional, as in modern recommender systems. Conversely, more flexible Bayesian models often require costly posterior approximations (e.g., nested Monte Carlo or variational inference), which can be difficult to deploy in real-time interactive settings. These challenges are amplified in \emph{cold-start} regimes \citep{schein2002methods}, where either the user is new (little or no history) and/or items/questions are new (limited calibration data), precisely when strong and structured priors are most valuable.

Recent advances in large language models (LLMs) suggest a new ingredient: LLMs can simulate plausible human responses when conditioned on rich textual profiles or personas, reproducing response patterns of specific demographic and attitudinal subgroups \citep{aher2023using, argyle2023out, horton2023large}. This capability has spurred interest in using personas for elicitation and adaptive questioning, but most existing approaches treat personas and LLM outputs as heuristic tools rather than as components of a coherent Bayesian model with principled posterior updates and decision-theoretic query selection. This motivates our central question: 
\begin{center}
\emph{Can we use AI personas to define a simple yet expressive Bayesian prior \\ that supports efficient adaptive querying?}
\end{center}

\begin{figure*}[ht]
    \centering
    \includegraphics[width=\linewidth]{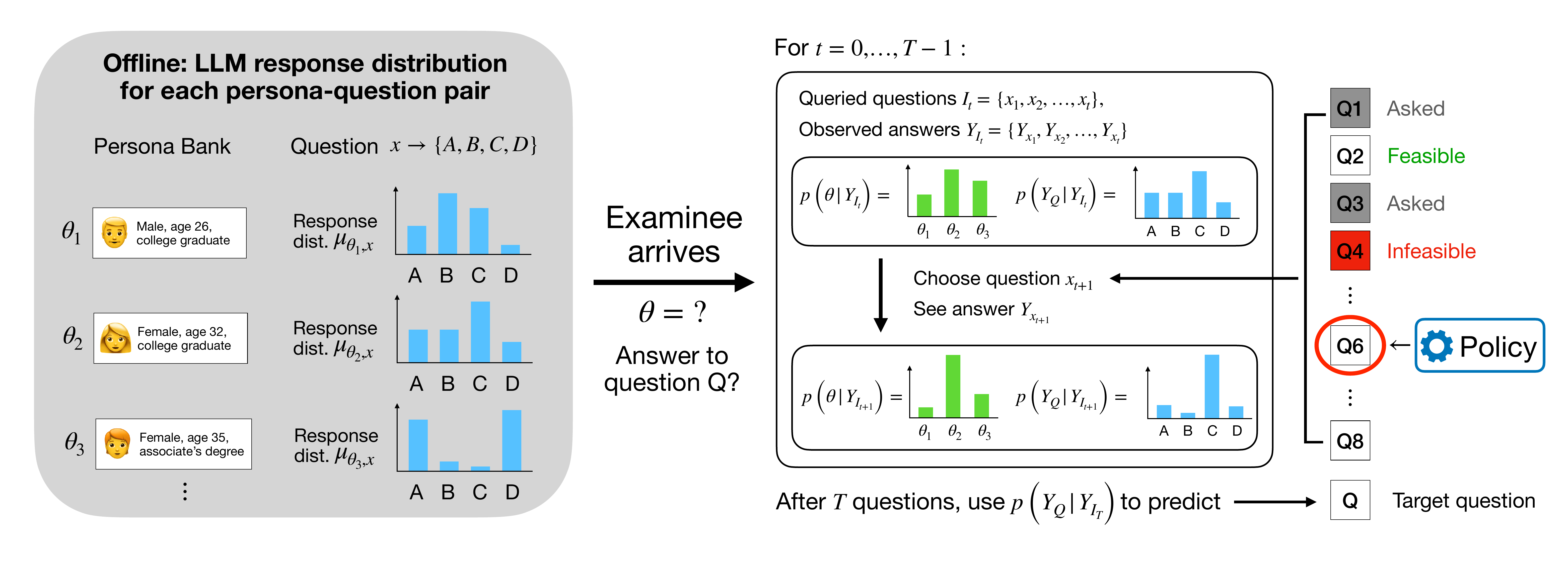}
    \caption{Workflow of our persona-based Bayesian adaptive querying. Offline, we collect persona--question response distributions from an LLM for a dictionary of personas. Online, a new user is modeled via a prior over persona membership, which is updated through Bayesian adaptive querying to form posterior beliefs and predictions. After exhausting a budget of questions, we make a probabilistic prediction on the user's answer to a target question.}
    \label{fig:workflow}
\end{figure*}

\paragraph{Overview.}
We propose \emph{Adaptive Querying with AI Persona Priors}. The key idea is to represent user heterogeneity through membership in a finite dictionary of AI personas, where each persona induces a distribution over responses to each question. We obtain these persona--question response distributions \emph{offline} by prompting an LLM; online, for a new user, we initialize a prior over persona membership and update it sequentially as answers arrive. The resulting posterior is a finite mixture with closed-form updates and predictions, enabling Bayesian experimental design (BED) methods and adaptive querying policies to be implemented efficiently. We illustrate the workflow in Figure~\ref{fig:workflow}.

\paragraph{Contributions.}
Our primary contribution is an \emph{end-to-end recipe} for turning LLM persona outputs into a Bayesian prior that supports closed-form posterior updates and efficient adaptive querying in noisy cold-start settings. Unlike classical CAT/IRT, our persona dictionary and response distributions are obtained entirely offline from an LLM, eliminating the need for task-specific item calibration and making the method immediately deployable. When training users are available, the prior over personas can also be adapted via empirical Bayes. Unlike recent neural BED approaches \citep{foster2021deep, ivanova2021implicit}, our model retains exact posterior inference, avoiding learned surrogates or policy networks. Concretely, our contributions include:
\begin{itemize}
    \item \textbf{A principled persona prior.} We introduce a \emph{persona-induced latent variable model} in which a user is represented by a member of a finite persona dictionary, with persona--question likelihoods provided by an LLM. This yields a simple but expressive prior over high-dimensional categorical response vectors.
    \item \textbf{Tractable Bayesian inference at scale.} Under categorical questions, the model admits \emph{closed-form posterior updates} over persona membership and \emph{finite-mixture posterior predictions}, avoiding nested Monte Carlo or variational approximations that commonly bottleneck BED in high dimensions.
    \item \textbf{Instantiation of classical adaptive methods.} We demonstrate how standard non-adaptive and adaptive Bayesian design strategies can be implemented efficiently within our persona-based Bayesian model, and note how RL-style formulations fit naturally into the same framework.
    \item \textbf{Empirical study with CAT as a reference point.} On synthetic data and WorldValuesBench, we evaluate persona-based posteriors as probabilistic predictors and compare against classical CAT/IRT baselines. The comparisons illustrate when persona priors can be especially effective, including cold-start user/item regimes that are challenging for calibration-heavy models.
\end{itemize}
The remainder of the paper is organized as follows. Section~\ref{sec:formulation} formalizes the adaptive querying problem; Section~\ref{sec:model} introduces our persona-induced latent variable model and inferential methodology; Section~\ref{sec:experiments} presents experiments on synthetic and real data; and Section~\ref{sec:related} relates our work to existing ones in related fields. We conclude the paper with a discussion in Section~\ref{sec:discussion}.

\section{Problem Formulation}\label{sec:formulation}

We study the problem of sequentially querying a user in order to learn about an unknown quantity of interest under a limited budget of questions. Let $\Y \in \mathcal Y^m$ denote a random vector representing the user's responses to a fixed bank of $m$ questions, where $\mathcal Y$ is a response space. User responses are assumed to be intrinsically noisy, and we model $\Y$ as a random vector drawn from a prior distribution $p_{\Y}$. Our objective is to infer a target quantity
\[
Z \;\triangleq\; g(\Y),
\]
where $g$ is a known mapping. The quantity $Z$ may represent, for example, the user's responses to a subset of unasked questions, a latent categorical label, a real-valued score, or a Bayes-optimal decision derived from $\Y$. In many applications, not all questions can be asked due to cost, sensitivity, or operational constraints. We model this by letting $\mathcal I_{\text{feas}} \subseteq [m]$ denote the set of feasible question indices.

\subsection{Setup}

\paragraph{Uncertainty-based objective.}
When the prior distribution $p_{\Y}$ is known, it induces a well-defined distribution over the target quantity $Z$. In this case, learning about $Z$ can be formalized as reducing uncertainty in its posterior distribution. Let $U(\cdot)$ denote a real-valued functional that measures uncertainty, such as Shannon entropy, variance, or Gini impurity. With slight abuse of notation, we write $U(X)$ to denote the uncertainty of a random variable $X$ through its distribution. The goal of adaptive querying is to design a policy that minimizes the uncertainty of $Z$ after a limited number of queries.

\paragraph{Bayesian adaptive querying.}
At each time step $t=1,\dots,T$ with $T \le m$, let
\[
h_t \;\triangleq\; (x_1,Y_{x_1},\dots,x_t,Y_{x_t})
\]
denote the interaction history, where $x_i \in \mathcal I_{\text{feas}}$ is the selected question and $Y_{x_i}$ is the corresponding observed response. We assume that each question can be asked at most once, and denote the set of queried questions by $I_t = \{x_1,\dots,x_t\}$. Define $h_0 = I_0 = \varnothing$. Conditioning on the history $h_t$, let
\[
P_t \;\triangleq\; p(Z \mid h_t)
\]
denote the posterior distribution of the target quantity. A Bayesian adaptive querying policy $\pi$ selects the next question $x_{t+1} \in \mathcal I_{\mathrm{feas}} \setminus I_t$ based on $h_t$, observes the response $Y_{x_{t+1}}$, and continues this process until the budget is exhausted. The performance of $\pi$ is measured by the uncertainty of the final posterior $P_T$. Formally, we seek to design a policy $\pi$ that minimizes the expected posterior uncertainty:
\begin{equation}\label{eq:problem}
\min_{\pi} \; \mathbb E\!\left[ U(P_T) \right],
\end{equation}
where the expectation is taken with respect to the randomness in the user's responses, induced jointly by the prior $p_{\Y}$ and the policy $\pi$. We may also write the objective as $E\!\left[ U(Z | h_T) \right]$.

\subsection{Evaluation and Scoring Rules}

In practice, the assumed prior $p_{\Y}$ is rarely exact, and posterior beliefs $P_t$ may be misspecified relative to the true data-generating process. This makes it essential to evaluate probabilistic predictions using statistically principled criteria.

We adopt the framework of \emph{proper scoring rules}. A scoring rule is a function $S(p,z)$ that assigns a numerical score to a predictive distribution $p$ when outcome $z$ is realized. It is \emph{strictly proper} if the expected score is uniquely maximized when $p$ coincides with the true distribution. Proper scoring rules therefore incentivize calibrated and honest probabilistic predictions.

A classical result establishes a close duality between uncertainty measures and proper scoring rules \citep{mccarthy1956measures, savage1971elicitation}. In particular, every strictly concave uncertainty functional $U$ induces a strictly proper scoring rule $S$, and conversely, any strictly proper scoring rule $S$ defines an uncertainty functional via its expected negative self-score,
\[
U_S(p) \;\triangleq\; -\mathbb E_{Z\sim p}[S(p,Z)].
\]
Canonical examples include Shannon entropy paired with logarithmic scoring and Gini impurity paired with the Brier score \citep{gneiting2007strictly}. This correspondence ensures that uncertainty-based query selection objectives align naturally with principled evaluation metrics, even under model misspecification.

\subsection{Approximate Solution Methods}

The optimization problem in \eqref{eq:problem} is combinatorial and generally NP-hard. Consequently, practical solutions rely on approximate methods \citep{rainforth2024modern}. We briefly outline several common approaches below; their efficient instantiations under our persona-based model are discussed in Section~\ref{sec:model}.

\paragraph{Non-adaptive optimal design.}
Classical Bayesian experimental design considers a non-adaptive setting in which all $T$ questions are selected upfront:
\begin{equation}\label{eq:nonadaptive_obj}
\min_{I \subseteq \mathcal I_{\mathrm{feas}}, |I|=T}
\; \mathbb E\!\left[ U(Z \mid Y_I) \right].
\end{equation}
This formulation avoids interaction-dependent computation and can be easier to deploy in practice. However, it ignores user-specific responses observed during querying and is therefore generally less sample-efficient. In practice, greedy forward selection heuristics---analogous to forward feature selection in regression---are commonly used to approximate \eqref{eq:nonadaptive_obj}.

\paragraph{Greedy adaptive querying.}
A widely used adaptive strategy is greedy one-step lookahead. At time $t$, for each candidate question $x \in \mathcal I_{\mathrm{feas}} \setminus I_t$, one computes the expected posterior uncertainty after observing its response,
\begin{equation}\label{eq:greedy-delta}
\Delta_U(x \mid h_t)
\;\triangleq\;
\mathbb E_{Y_x \sim p(\cdot \mid h_t)}
\Big[ U(Z \mid h_t, Y_x) \Big],
\end{equation}
and selects the question that minimizes this quantity. This procedure prioritizes questions with the largest expected immediate reduction in uncertainty. Extensions to multi-step lookahead or tree search are possible but are typically more computationally demanding.

\paragraph{Reinforcement learning.}
The scoring-rule perspective naturally yields a non-myopic reinforcement learning (RL) formulation of adaptive querying. The interaction between the agent and the user defines a finite-horizon episodic decision process, where actions correspond to question selections and observations correspond to responses. We can define the reward in step $t$ as $U(P_{t-1}) - U(P_{t})$, which measures the uncertainty reduction. The cumulative reward over $T$ steps is $U(P_0)-U(P_T)$, making the RL objective equivalent to minimizing final posterior uncertainty.

\vskip\baselineskip
Beyond these approaches, the formulation in \eqref{eq:problem} also encompasses Thompson sampling-style policies, Bayesian optimization acquisition functions, and other information-theoretic strategies. We leave a systematic comparison of these methods to future work.

\section{Methodology}\label{sec:model}

The adaptive querying strategies described in Section~\ref{sec:formulation} are agnostic to the choice of prior distribution $p_{\Y}$. In practice, however, their successful deployment hinges on the ability to efficiently compute posterior distributions and predictive likelihoods at each step of the interaction. For general high-dimensional priors with complex dependencies across questions, posterior inference can be intractable, rendering even greedy adaptive methods computationally prohibitive.

This motivates the use of structured probabilistic models that balance \emph{expressiveness}---the ability to capture rich and heterogeneous user response patterns---with \emph{tractability}---the ability to support fast posterior updates and prediction. In this section, we introduce a latent variable model based on AI personas that achieves this balance. The resulting model admits closed-form inference while leveraging LLMs to encode complex prior information.

\subsection{Persona-Induced Latent Variable Model}

A standard way to impose structure on $p_{\Y}$ is through a latent variable $\theta \in \Theta$ that captures user-specific characteristics. We assume conditional independence of responses across questions given $\theta$, yielding the joint model
\begin{equation}
p(\theta, \Y) \;=\; p_\theta(\theta) \prod_{i=1}^m p(Y_i \mid \theta).
\end{equation}
This is a simplifying assumption shared with classical IRT and CAT models that enables closed-form posterior updates. Given observations $\Y_{I_t}$, Bayes' rule gives the posterior
\begin{equation}\label{eq:posterior_theta}
p(\theta \mid \Y_{I_t}) \;\propto\; p_\theta(\theta) \prod_{i \in I_t} p(Y_i \mid \theta),
\end{equation}
and the posterior predictive distribution for an unasked question $x$ is
\begin{equation}\label{eq:posterior_predictive}
p(Y_x \mid \Y_{I_t})
\;=\;
\int p(Y_x \mid \theta)\, p(\theta \mid \Y_{I_t}) \, d\theta.
\end{equation}

This latent-variable formulation allows posterior predictive sampling via a two-step procedure: first sample $\theta$ from $p(\theta \mid \Y_{I_t})$, then sample $Y_x$ from $p(\cdot \mid \theta)$. However, in the context of Bayesian adaptive querying, the predictive integral in \eqref{eq:posterior_predictive} is typically nested inside expectations over future observations (cf.~\eqref{eq:problem}), leading to repeated high-dimensional integrations at every decision step.

This challenge is well known in Bayesian experimental design and has motivated approaches such as nested Monte Carlo estimation \citep{rainforth2018nesting} and variational approximations \citep{foster2019variational}. While effective in some settings, these methods remain computationally intensive and introduce accuracy-efficiency trade-offs. This motivates the search for a latent variable model that is both expressive and admits efficient posterior updates.

Recent advances in LLMs provide a compelling answer. LLMs can generate coherent, human-like responses when conditioned on descriptive profiles or personas, suggesting a natural way to encode rich prior beliefs about user behavior. Suppose we are given a dictionary of $n$ AI personas with profiles $\xi_1,\dots,\xi_n$. For each persona and question, we can query an LLM conditioned on the persona profile to obtain an estimated response distribution.

If the persona dictionary is sufficiently representative, it is reasonable to model a new user as one (or a mixture) of these personas. Accordingly, we define the latent variable as persona membership,
\[
\theta \in \{1,2,\dots,n\},
\]
and interpret the user as being drawn from persona $\theta$.\footnote{While the prior over $\theta$ corresponds to a hard assignment, the posterior is soft and supports mixture-like inference.} We posit a prior $p(\theta)$ and define the item-response model as
\[
\Y_x \mid \theta \;=\; \mathsf{LLM}(\xi_\theta, x).
\]
The response distribution $\mathsf{LLM}(\xi_\theta, x)$ can be obtained in various ways, including prompting, log-probability extraction, or calibrated sampling; see Appendix~\ref{app:response_dist} for details. This construction turns LLM-based personas into an explicit probabilistic prior rather than a heuristic simulation tool, enabling principled Bayesian inference.

\paragraph{Categorical questions.}
For clarity and concreteness, we focus on the setting where all questions have $K$ categorical responses, so that $\Y \in \{1,2,\dots,K\}^m$. For each persona-question pair, we model the response distribution as categorical with parameter
\[
\mu_{\theta,x} = (\mu_{\theta,x,1}, \dots, \mu_{\theta,x,K}) \in \Delta^{K-1},
\]
so that
\[
\Y_x \mid \theta
\;=\;
\mathsf{LLM}(\xi_\theta, x)
\;=\;
\text{Categorical}(\mu_{\theta,x}).
\]
Under this model, Bayesian inference admits closed-form expressions. The posterior over persona membership after observing $\Y_{I_t}$ is
\begin{equation}\label{eq:posterior_theta_categorical}
p(\theta \mid \Y_{I_t})
\;\propto\;
p(\theta)\prod_{i \in I_t} \mu_{\theta,i,\Y_i},
\end{equation}
which can be normalized efficiently since $\theta$ ranges over a finite set. The posterior predictive distribution for an unasked question $x$ is then
\begin{equation}\label{eq:posterior_predictive_categorical}
p(Y_x = k \mid \Y_{I_t})
\;=\;
\sum_{\theta=1}^n
\mu_{\theta,x,k}\, p(\theta \mid \Y_{I_t}).
\end{equation}

\paragraph{Discussion.}
This finite-mixture structure combines the expressiveness of LLM-generated response distributions with the computational simplicity of discrete latent variable models. Structurally, the persona-induced model is a finite mixture model, connecting it to latent class analysis \citep{goodman1974exploratory}. However, it differs in that the mixture components are not estimated from task-specific data but are instead defined by LLM-elicited response distributions. Moreover, the latent variable $\theta$ has a clear semantic interpretation as persona membership, enabling downstream tasks such as user clustering, response simulation, and group-level analysis. Importantly, the framework is model-agnostic and can be instantiated with \emph{any} pre-trained or fine-tuned LLM.

\subsection{Non-Adaptive Optimal Design}

We first consider non-adaptive Bayesian optimal design under the persona-induced model. In this setting, all $T$ questions are selected \emph{a priori} before observing any responses, corresponding to the batch formulation of Bayesian experimental design. For a candidate set $I$, the expected posterior uncertainty can be written as
\begin{equation}\label{eq:nonadaptive_expected}
\mathbb E\!\left[ U(Z \mid \Y_I) \right]
=
\sum_{y_I \in \mathcal Y^{|I|}}
p(\Y_I = y_I)\,
U\!\left(Z \mid \Y_I = y_I\right),
\end{equation}
where the marginal likelihood is
\[
p(\Y_I = y_I)
=
\sum_{\theta=1}^n
p(\theta)\prod_{i \in I} \mu_{\theta,i,y_i}.
\]
Although \eqref{eq:nonadaptive_expected} is available in closed form, selecting the optimal subset in \eqref{eq:nonadaptive_obj} remains a combinatorial optimization problem and is generally NP-hard. A common approximation is greedy forward selection: starting from $I_0=\varnothing$, at each step select
\begin{equation}\label{eq:nonadaptive_greedy}
x_{t+1}
\;\in\;
\argmin_{x \in \mathcal I_{\mathrm{feas}} \setminus I_t}
\mathbb E\!\left[
U\big(Z \mid \Y_{I_t \cup \{x\}}\big)
\right].
\end{equation}
Unlike adaptive querying, this expectation is computed before any responses are observed, and the resulting question set is fixed across users. Algorithm~\ref{alg:nonadaptive_bed} summarizes this procedure.

\begin{algorithm}[ht]
  \caption{Greedy Non-Adaptive Bayesian Optimal Design}
  \label{alg:nonadaptive_bed}
  \begin{algorithmic}[1]
    \Require
    Budget $T$; feasible questions $\mathcal I_{\mathrm{feas}}$;
    prior $p(\theta)$; likelihoods $\{\mu_{\theta,x}\}$;
    uncertainty functional $U(\cdot)$
    \State Initialize $I_0 \gets \varnothing$
    \For{$t = 0,1,\dots,T-1$}
        \ForAll{$x \in \mathcal I_{\mathrm{feas}} \setminus I_t$}
            \State Compute expected posterior uncertainty
            \[
            \Delta_U^{\mathrm{batch}}(x \mid I_t)
            \;=\;
            \mathbb E\!\left[
            U\big(Z \mid \Y_{I_t \cup \{x\}}\big)
            \right]
            \]
            using \eqref{eq:nonadaptive_expected}
        \EndFor
        \State Select
        $
        x_{t+1}
        \gets
        \argmin_{x \in \mathcal I_{\mathrm{feas}} \setminus I_t}
        \Delta_U^{\mathrm{batch}}(x \mid I_t)
        $
        \State Update $I_{t+1} \gets I_t \cup \{x_{t+1}\}$
    \EndFor
    \State {\bfseries Return:} Fixed question set $I_T$
  \end{algorithmic}
\end{algorithm}

Non-adaptive designs are simple to deploy and avoid online interactive computation. However, they cannot tailor queries to individual users and are therefore typically less sample-efficient than adaptive methods. At the same time, they may be more robust to model misspecification and overly aggressive adaptivity.

\subsection{Greedy Adaptive Querying}

Under the categorical persona model, greedy adaptive querying from Section~\ref{sec:formulation} becomes particularly efficient. The one-step lookahead objective in \eqref{eq:greedy-delta} reduces to
\begin{equation}\label{eq:greedy-delta-categorical}
\Delta_U(x \mid h_t)
=
\sum_{k=1}^K
p(Y_x = k \mid \Y_{I_t})\,
U\big(Z \mid h_t, Y_x = k\big),
\end{equation}
where each term is computed using \eqref{eq:posterior_theta_categorical} and \eqref{eq:posterior_predictive_categorical}. Algorithm~\ref{alg:greedy_baq} summarizes the resulting greedy Bayesian adaptive querying procedure.

\begin{algorithm}[ht]
  \caption{Greedy Bayesian Adaptive Query}
  \label{alg:greedy_baq}
  \begin{algorithmic}[1]
    \Require Budget $T$; feasible questions $\mathcal I_{\mathrm{feas}}$;
    prior $p(\theta)$; likelihoods $\{\mu_{\theta,x}\}$;
    uncertainty functional $U(\cdot)$
    \State $I_0\gets\varnothing$, $\Y_{I_0}\gets \varnothing$
    \For{$t = 0,1,\dots,T-1$}
        \ForAll{$x \in \mathcal I_{\mathrm{feas}} \setminus I_t$}
            \State Compute $p(Y_x \mid \Y_{I_t})$ using \eqref{eq:posterior_theta_categorical} and \eqref{eq:posterior_predictive_categorical}
            \State Compute $\Delta_U(x \mid \Y_{I_t})$ using \eqref{eq:greedy-delta-categorical}
        \EndFor
        \State Select $x_{t+1} \gets \argmin_{x \in \mathcal I_{\mathrm{feas}} \setminus I_t} \Delta_U(x \mid \Y_{I_t})$
        \State Query question $x_{t+1}$ and observe answer $Y_{x_{t+1}}$
        \State Update $I_{t+1} \gets I_t \cup \{x_{t+1}\}$, $\Y_{I_{t+1}} \gets (\Y_{I_t},\Y_{x_{t+1}})$
    \EndFor
    \State {\bfseries Return:} Observed answers $\Y_{I_T}$
  \end{algorithmic}
\end{algorithm}

\paragraph{Connection to collaborative filtering (CF).}
Our approach bears a resemblance to CF and lookalike modeling, which also leverage population-level patterns to predict individual preferences \citep{su2009survey}. However, they differ in several important respects. First, our model is a \emph{generative Bayesian model} with an explicit latent variable and closed-form posterior updates, rather than a similarity-based or matrix-factorization approach. Second, our persona prior requires \emph{no historical response data} from the target population---it is constructed entirely from LLM-generated persona profiles, making it suitable for cold-start settings where CF methods have insufficient data. Third, our framework supports \emph{decision-theoretic query selection}, actively choosing which questions to ask to reduce posterior uncertainty, rather than passively processing available ratings.

\section{Experiments}\label{sec:experiments}

We evaluate the proposed persona-based Bayesian adaptive querying framework on synthetic users and on real users from WorldValuesBench, with classical CAT methods as reference baselines.

\paragraph{Implementation overview.}
We use GPT-5-mini for all persona-conditioned response distribution elicitation (Appendix~\ref{app:response_dist}--\ref{app:llm_prompts}), the Twin-2K-500 persona bank \citep{toubia2025database} as the persona dictionary (Section~\ref{sec:datasets}), and custom implementations of polytomous CAT baselines (Appendix~\ref{app:cat_details}). Ablation studies are reported in Section~\ref{sec:ablations}.

\subsection{Datasets and Persona Construction}\label{sec:datasets}

\paragraph{WorldValuesBench \citep{zhao2024worldvaluesbench}.}
WorldValuesBench contains survey responses from over 94{,}000 participants to 290 questions on values and beliefs (e.g., family, politics, religion, work, and society). We restrict attention to ordinal Likert-style questions with four categories and filter out respondents with more than 20\% missing answers. The resulting dataset contains 91 questions and 88{,}459 users, with an overall missing rate of 2.6\%.

\paragraph{Handling missing responses.}
For a given user, if the response to a question is missing, we treat that question as infeasible for that user (i.e., it cannot be queried). During evaluation, metrics are computed only on user--question pairs with observed ground-truth responses. This convention ensures that users with more missing data naturally retain higher posterior uncertainty, since fewer observations are available to update their latent membership.

\paragraph{Persona dictionary and response distributions.}
We use the Twin-2K-500 persona bank \citep{toubia2025database} as a dictionary of $n = 2{,}058$ latent profiles. Each persona corresponds to a real U.S.\ participant whose responses to over 500 questions spanning demographic, psychological, economic, and behavioral domains have been collected. The dictionary has three desirable properties: (i) \emph{diversity}---personas cover a broad range of demographic backgrounds and attitudinal profiles; (ii) \emph{domain coverage}---the underlying question bank spans topics well beyond WorldValuesBench, providing rich conditioning information; and (iii) \emph{grounding}---each persona is anchored to a real individual's response pattern, reducing the risk of generating unrealistic or incoherent profiles. These personas do not correspond to real users in WorldValuesBench; instead, they provide a structured, interpretable prior over response patterns. For each persona $\xi_\theta$ and each question $x$, we prompt an LLM (GPT-5-mini) to produce a categorical distribution over the four Likert responses, yielding parameters $\mu_{\theta,x} \in \Delta^{3}$. Appendices~\ref{app:response_dist} and \ref{app:llm_prompts} detail the elicitation strategy and the prompts.

\paragraph{Synthetic users (well-specified prior).}
To study behavior under correct specification, we generate synthetic users from the persona model. For each synthetic user $j$, we sample a single persona $\theta^{(j)} \sim p(\theta)$ and then sample responses as
\[
Y^{(j)}_x \sim \mathrm{Categorical}(\mu_{\theta^{(j)},x})
\]
for each question $x$.

\subsection{Experimental Protocol and Evaluation}

We split users into training (80\%) and test (20\%) sets. All evaluations are performed on held-out test users. For each test user, the algorithm sequentially selects questions from a feasible set and observes the corresponding ground-truth responses; after a budget of $T$ queries, we evaluate the resulting posterior predictive distribution on target questions.

\paragraph{Target questions and feasible set.}
We consider a held-out prediction task in which a small subset of questions $I^\star$ are designated as \emph{targets}, i.e., $g(\Y)=\Y_{I^\star}$, while the remaining questions constitute the feasible set $\mathcal I_{\mathrm{feas}}$. This setting captures applications where a small set of key indicators is of primary interest and must be inferred from a limited interactive budget. In our experiments, we randomly select 5 questions as targets (fixed across test users), leaving the remaining 86 questions as the feasible set.

\paragraph{Metrics.}
Let $\hat p_{u,q}$ denote the predictive distribution for user $u$ on target question $q \in I^\star$, and let $y_{u,q}$ denote the realized response. We report:
\begin{itemize}
    \item \textbf{Log loss:} $-\log \hat p_{u,q}(y_{u,q})$.
    \item \textbf{Brier score:} $\sum_{k=1}^K (\hat p_{u,q}(k) - \mathbf{1}\{y_{u,q}=k\})^2$.
    \item \textbf{Ordinal MSE:} squared error between the posterior mean under $\hat p_{u,q}$ and the ordinal-coded outcome (categories mapped to $\{0,1,2,3\}$).
\end{itemize}
Metrics are averaged over all $(u,q)$ pairs in the test set.

\subsection{Methods Compared}

\paragraph{Persona-based querying policies.}
Unless otherwise stated, persona-based methods use as uncertainty functional the sum of Shannon entropies of the target marginals, $U(P_t)=\sum_{x' \in I^\star} H(Y_{x'} \mid h_t)$.\footnote{Thus, for the vector-valued target $Z=\Y_{I^\star}$ in the held-out task, we evaluate uncertainty additively across target coordinates rather than through the joint entropy of $Z$. The joint entropy is tractable but adds over exponentially many terms in the number of target questions, while the sum of marginal entropies is more scalable and still captures the overall uncertainty in the target.} In the held-out task, the one-step lookahead objective becomes
\[
\Delta_U(x \mid h_t)
= \sum_{k=1}^K p(Y_x=k \mid \Y_{I_t})
\sum_{x' \in I^\star}
H\!\left(\Y_{x'} \mid h_t, Y_x=k\right),
\]
i.e., the expected posterior sum of marginal target entropies after querying $x$.

\paragraph{Prior specification.}
For experiments on real users, we learn the prior $p(\theta)$ from training users via empirical Bayes by maximizing the marginal likelihood
\[
\max_{p(\theta) \in \Delta^{n-1}}
\sum_{j=1}^N
\log\!\left(
\sum_{\theta}
p(\theta)\, p(\Y^{(j)} \mid \theta)
\right).
\]
We optimize this objective with an EM algorithm: the E-step computes responsibilities $\gamma_{j,\theta} \propto p(\theta)\, p(\Y^{(j)} \mid \theta)$, and the M-step updates $p(\theta) = \frac{1}{N}\sum_{j=1}^N \gamma_{j,\theta}$. The learned prior downweights personas that are rarely matched to real users, concentrating mass on the most relevant region of persona space and mitigating the misspecification inherent in applying a synthetic persona dictionary to a real population. For synthetic users, where data is generated from the persona model, we use a uniform prior.

In addition to Algorithm~\ref{alg:nonadaptive_bed} and Algorithm~\ref{alg:greedy_baq}, we compare several additional persona-based baselines. The \textbf{Random} strategy selects feasible questions uniformly at random at each step, while \textbf{Random Fixed} selects a fixed set of $T$ questions uniformly at random for all users. We also include a \textbf{Full} baseline that queries all feasible questions; this serves as an oracle upper bound on available information, though not necessarily on predictive performance when the persona prior is misspecified.

\paragraph{CAT baselines.}
We implement classical polytomous CAT methods based on item response theory (IRT). Specifically, we consider the graded response model (GRM) and generalized partial credit model (GPCM), each in both one-dimensional and multidimensional variants (MGRM/MGPCM) \citep{samejima1969estimation, muraki1992generalized, wainer2000computerized, yao2006multidimensional, reckase2009multidimensional, van2010elements}. For each model, we fit item parameters on the training users via marginal maximum likelihood (EM) and perform inference with a grid-based posterior over latent traits. Since existing open-source CAT libraries do not robustly support multidimensional polytomous settings, we implement these baselines from scratch; details are in Appendix~\ref{app:cat:implementation}.

The two families of methods differ sharply in what they require from training data. CAT baselines must calibrate item parameters (discriminations and thresholds) for \emph{every item in the bank}, which requires a large number of user responses to each item. If new items are introduced or the item bank changes, recalibration is necessary. In contrast, persona-based methods obtain item-level response distributions entirely from the LLM---no observed responses to those items are needed. The only component learned from training data is the prior $p(\theta)$ over personas, a single $n$-dimensional weight vector that does not depend on the identity of individual items.

In our experiments, we provide CAT with more than 70{,}000 training users---sufficient for reliable item calibration---making this a generous test of CAT performance. Persona-based methods use the same training split, but only to fit the persona prior. The goal of this setup is to show that, even under favorable conditions for CAT, persona-based methods remain competitive with or superior to this well-established and effective approach. When calibration data is scarce or unavailable---as in cold-start item regimes where new questions must be deployed without prior response data---CAT simply cannot be applied, whereas persona-based methods can incorporate new items immediately via LLM prompting.

\subsection{Results}

\subsubsection{Synthetic users (well-specified model)}
\begin{figure*}[t]
\centering
\begin{subfigure}{.5\textwidth}
\centering
\includegraphics[width=.95\linewidth]{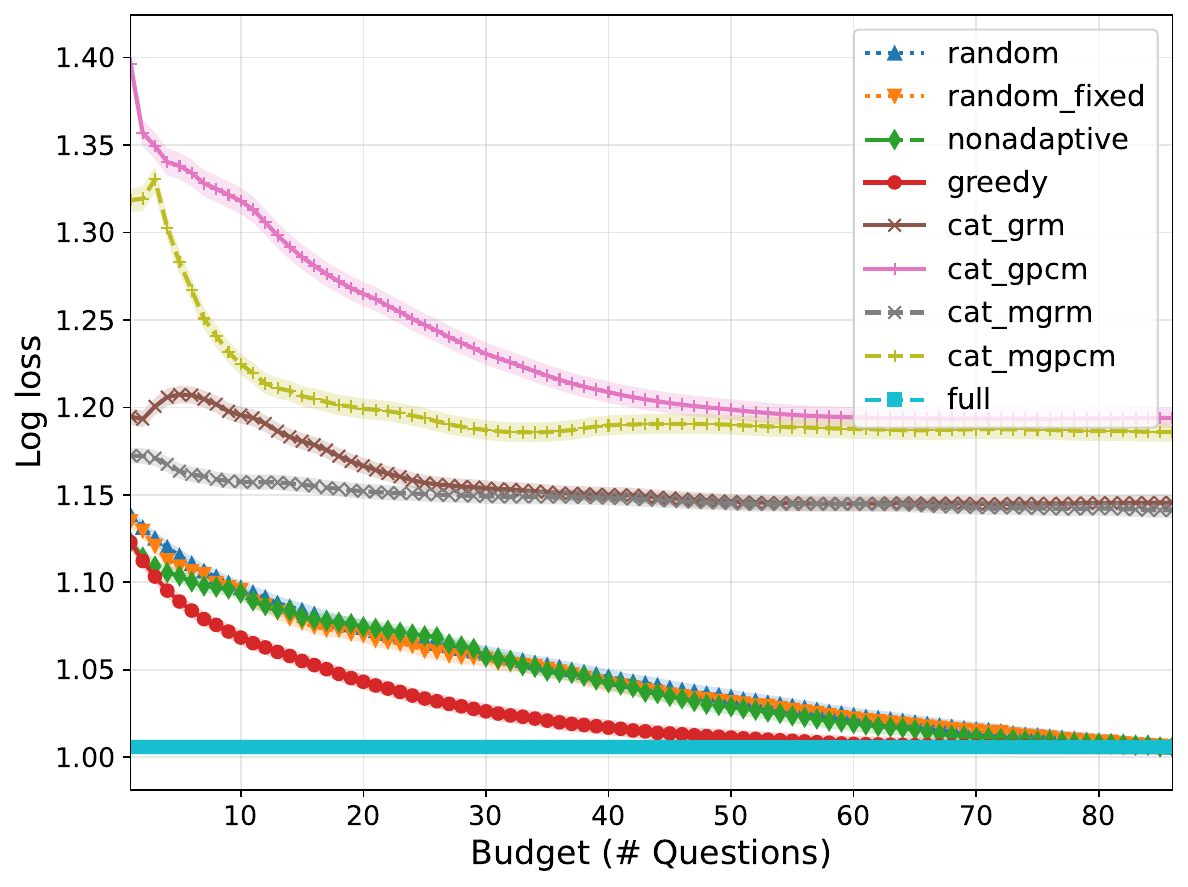}
\caption{Synthetic users (well-specified prior).}
\label{fig:synthetic_results_log}
\end{subfigure}%
\begin{subfigure}{.5\textwidth}
\centering
\includegraphics[width=.95\linewidth]{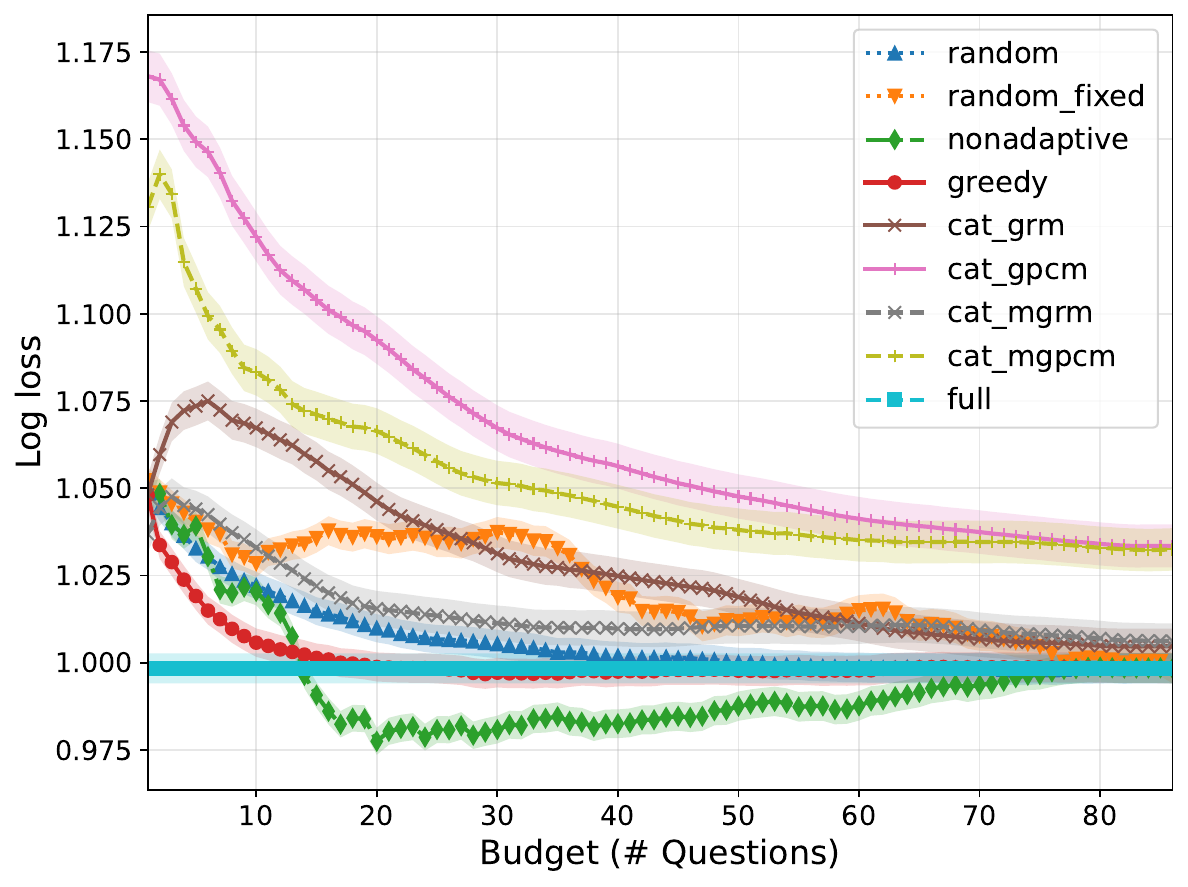}
\caption{Real users (WorldValuesBench).}
\label{fig:real_results_log}
\end{subfigure}
\caption{Log loss versus query budget. Curves denote mean log loss averaged over all user--target-question pairs; shaded regions indicate 95\% confidence intervals. Left: when the persona prior is correctly specified, persona-based methods substantially outperform CAT baselines, with greedy achieving the fastest error reduction. Right: under model misspecification on real data, persona-based methods still outperform CAT; greedy performs best at small budgets, while non-adaptive designs can overtake at larger budgets.}
\label{fig:main_results_log}
\end{figure*}

We first evaluate on 100{,}000 synthetic users sampled from the persona model, where the prior is correctly specified. Figure~\ref{fig:synthetic_results_log} and Table~\ref{tab:synthetic_logloss_main} report log loss versus query budget. As expected, persona-based methods dominate CAT baselines in the well-specified setting: the persona model matches the data-generating process, while IRT-based CAT is structurally misspecified. In the left panel of Figure~\ref{fig:main_results_log}, performance improves approximately monotonically with budget for all persona-based methods, and the \textbf{Full} curve serves as an upper bound, suggesting that with synthetic data all questions are informative about the target questions. Greedy achieves the fastest reduction in log loss, confirming it as a strong and simple adaptive heuristic. Additional metrics (Brier score and ordinal MSE) show the same qualitative behavior (Appendix~\ref{app:additional_world}).

\begin{table}[ht]
\caption{Synthetic users: log loss by query budget $T$. $N = 20{,}000$ test users; cells report mean with standard error below. At $T = 86$ all feasible questions have been asked, so all persona-based methods coincide with the \textbf{full} baseline. \textbf{Bold} marks the best value per column.}
\label{tab:synthetic_logloss_main}
\centering
\small
\begin{tabular}{l*{7}{c}}
\toprule
\textbf{Method} & $T\!=\!5$ & $T\!=\!10$ & $T\!=\!15$ & $T\!=\!20$ & $T\!=\!30$ & $T\!=\!50$ & $T\!=\!86$ \\
\midrule
random        & \cell{1.115}{.002} & \cell{1.097}{.002} & \cell{1.084}{.002} & \cell{1.074}{.002} & \cell{1.059}{.002} & \cell{1.034}{.002} & \cell{\textbf{1.006}}{.002} \\[4pt]
random\_fixed & \cell{1.109}{.002} & \cell{1.095}{.002} & \cell{1.077}{.002} & \cell{1.070}{.002} & \cell{1.056}{.002} & \cell{1.031}{.002} & \cell{\textbf{1.006}}{.002} \\[4pt]
nonadaptive   & \cell{1.104}{.002} & \cell{1.094}{.002} & \cell{1.080}{.002} & \cell{1.074}{.002} & \cell{1.057}{.002} & \cell{1.029}{.002} & \cell{\textbf{1.006}}{.002} \\[4pt]
greedy        & \cell{\textbf{1.089}}{.002} & \cell{\textbf{1.068}}{.002} & \cell{\textbf{1.055}}{.002} & \cell{\textbf{1.043}}{.002} & \cell{\textbf{1.026}}{.002} & \cell{\textbf{1.011}}{.002} & \cell{\textbf{1.006}}{.002} \\[4pt]
\midrule
CAT-GRM       & \cell{1.207}{.003} & \cell{1.196}{.002} & \cell{1.181}{.002} & \cell{1.167}{.002} & \cell{1.154}{.002} & \cell{1.146}{.002} & \cell{1.146}{.002} \\[4pt]
CAT-GPCM      & \cell{1.338}{.004} & \cell{1.318}{.004} & \cell{1.286}{.004} & \cell{1.265}{.004} & \cell{1.231}{.003} & \cell{1.199}{.003} & \cell{1.194}{.003} \\[4pt]
CAT-MGRM      & \cell{1.164}{.002} & \cell{1.157}{.002} & \cell{1.156}{.002} & \cell{1.152}{.002} & \cell{1.149}{.002} & \cell{1.145}{.002} & \cell{1.141}{.002} \\[4pt]
CAT-MGPCM     & \cell{1.283}{.003} & \cell{1.225}{.003} & \cell{1.206}{.003} & \cell{1.199}{.003} & \cell{1.187}{.003} & \cell{1.190}{.003} & \cell{1.186}{.003} \\
\bottomrule
\end{tabular}
\end{table}

\subsubsection{Real users (misspecified model)}
We next evaluate on held-out WorldValuesBench users (Figure~\ref{fig:real_results_log} and Table~\ref{tab:real_logloss}). Compared to synthetic users, gains over CAT persist but are smaller, consistent with inevitable model misspecification on real data. Notably, the \textbf{Full} curve in Figure~\ref{fig:real_results_log} does not always yield the best predictive performance, suggesting that additional queried questions can sometimes be weakly informative or even misleading for predicting the target set under a misspecified model.

\begin{table}[ht]
\caption{Real users (WorldValuesBench): log loss by query budget $T$ for all methods. $N = 17{,}692$ held-out users; cells report mean with standard error below. At $T = 86$ all feasible questions have been asked, so all persona-based methods coincide with the \textbf{full} baseline. \textbf{Bold} marks the best value per column.}
\label{tab:real_logloss}
\centering
\small
\begin{tabular}{l*{7}{c}}
\toprule
\textbf{Method} & $T\!=\!5$ & $T\!=\!10$ & $T\!=\!15$ & $T\!=\!20$ & $T\!=\!30$ & $T\!=\!50$ & $T\!=\!86$ \\
\midrule
random        & \cell{1.033}{.002} & \cell{1.022}{.002} & \cell{1.015}{.002} & \cell{1.010}{.002} & \cell{1.005}{.002} & \cell{1.000}{.002} & \cell{\textbf{.998}}{.002} \\[4pt]
random\_fixed & \cell{1.040}{.002} & \cell{1.028}{.002} & \cell{1.036}{.002} & \cell{1.036}{.002} & \cell{1.037}{.002} & \cell{1.012}{.002} & \cell{\textbf{.998}}{.002} \\[4pt]
nonadaptive   & \cell{1.039}{.002} & \cell{1.020}{.002} & \cell{\textbf{.991}}{.002} & \cell{\textbf{.977}}{.002} & \cell{\textbf{.981}}{.002} & \cell{\textbf{.988}}{.002} & \cell{\textbf{.998}}{.002} \\[4pt]
greedy        & \cell{\textbf{1.019}}{.002} & \cell{\textbf{1.006}}{.002} & \cell{1.001}{.002} & \cell{.999}{.002} & \cell{.997}{.002} & \cell{.998}{.002} & \cell{\textbf{.998}}{.002} \\[4pt]
\midrule
CAT-GRM       & \cell{1.074}{.003} & \cell{1.067}{.003} & \cell{1.058}{.003} & \cell{1.046}{.003} & \cell{1.031}{.003} & \cell{1.019}{.003} & \cell{1.005}{.003} \\[4pt]
CAT-GPCM      & \cell{1.149}{.004} & \cell{1.122}{.004} & \cell{1.104}{.004} & \cell{1.092}{.003} & \cell{1.067}{.003} & \cell{1.048}{.003} & \cell{1.033}{.003} \\[4pt]
CAT-MGRM      & \cell{1.044}{.003} & \cell{1.033}{.003} & \cell{1.022}{.003} & \cell{1.015}{.003} & \cell{1.011}{.003} & \cell{1.011}{.003} & \cell{1.006}{.003} \\[4pt]
CAT-MGPCM     & \cell{1.107}{.004} & \cell{1.083}{.003} & \cell{1.071}{.003} & \cell{1.066}{.003} & \cell{1.051}{.003} & \cell{1.038}{.003} & \cell{1.032}{.003} \\
\bottomrule
\end{tabular}
\end{table}

A striking pattern is that greedy performs best for small budgets (e.g., $T \le 15$ in our experiments) but can be overtaken by the non-adaptive design at larger budgets. Moreover, beyond a moderate budget, the non-adaptive curve can degrade as $T$ increases, reinforcing that under misspecification, ``more questions'' need not imply better held-out predictions. We view this as evidence that short-horizon adaptivity can overfit to locally informative queries when the prior is imperfect, and that robust batch designs may sometimes provide better long-run prediction. Brier score and ordinal MSE results are consistent and reported in Appendix~\ref{app:additional_world}.

\paragraph{Efficiency-robustness tradeoff.}
Under correct specification, greedy one-step lookahead is near-optimal for entropy-based uncertainty-reduction objectives. Under misspecification, however, greedy querying can \emph{overcommit}: by selecting questions that are maximally informative under the current (misspecified) posterior, it may narrow the posterior prematurely onto an incorrect region of persona space. Subsequent questions are then chosen to refine an already-biased posterior, producing a cascade of locally optimal but globally suboptimal selections. In contrast, non-adaptive designs select a diverse, user-independent question set that hedges against misspecification by not conditioning on potentially misleading intermediate observations. This explains the crossover observed in Figure~\ref{fig:real_results_log}: greedy dominates at small budgets where its uncertainty-reduction advantage outweighs misspecification effects, while non-adaptive designs become more robust at larger budgets where greedy's accumulated bias degrades predictions.

\subsection{Ablation Studies}\label{sec:ablations}

We probe the robustness of persona priors along three axes that target the most plausible failure modes: \emph{dictionary granularity} (does using fewer personas hurt?), \emph{distributional richness} (does the full LLM-elicited distribution carry information beyond its mode?), and \emph{calibration} (are the LLM-elicited probabilities already well-tuned for the adaptive querying objective?). Tables~\ref{tab:ablation_nonadaptive} and~\ref{tab:ablation_greedy} summarize log-loss results for the \textbf{nonadaptive} and \textbf{greedy} methods, respectively, across these three axes. Brier score and ordinal MSE show the same qualitative patterns.

\paragraph{Persona dictionary clustering.}
The full Twin-2K-500 dictionary contains $n = 2{,}058$ personas, which may be larger than necessary for effective inference. To assess sensitivity to dictionary size, we compress the dictionary into a smaller set of prototype personas. Concretely, we first prune low-mass personas using the empirical Bayes prior learned from training users, then cluster the remaining personas with prior-weighted $k$-means using Jensen--Shannon divergence as the distance metric, and construct each prototype as the prior-weighted average of member personas' response distributions across questions. The prior over prototypes is set to the sum of priors of personas assigned to each cluster, preserving Bayesian consistency in the reduced dictionary. Tables~\ref{tab:ablation_nonadaptive} and~\ref{tab:ablation_greedy} show results for $n \in \{50, 200\}$ clusters for non-adaptive and greedy methods, respectively. Performance is robust down to approximately 200 clusters, with limited degradation at 50 clusters. At 200 clusters, the greedy method even slightly improves at some budgets, likely because pruning redundant or noisy personas reduces posterior diffusion. This suggests that moderate compression can reduce computational cost with minimal loss in predictive accuracy.

\paragraph{Deterministic-with-noise responses.}
A natural question is whether the full distributional shape of the LLM-elicited response probabilities matters, or whether a simpler point-prediction approach suffices. To test this, we replace the elicited distributions with a deterministic (mode) response plus uniform noise: for each persona--question pair, we first prompt the LLM to output a single canonical answer $\hat{y}$ (the most likely response option), and then define a noisy categorical distribution $p(y) = (1-\varepsilon)\mathbf{1}\{y = \hat{y}\} + \varepsilon/(K-1)$, where $\varepsilon \in \{0.1, 0.3\}$ controls the sharpness of the persona model. This ablation uniformly degrades performance across all methods and budgets---often dramatically so (e.g., log loss exceeding 1.3 at moderate budgets for $\varepsilon = 0.1$). The degradation is especially severe for small $\varepsilon$, where the near-deterministic likelihoods cause the posterior to concentrate rapidly on a single persona, leaving little room for correction after early misassignment. This confirms that the distributional shape of the LLM-elicited responses carries substantial information beyond the modal answer, and that directly eliciting probability distributions from the LLM is a meaningfully better strategy than eliciting point predictions and injecting synthetic noise.

\paragraph{Temperature scaling.}
We apply temperature scaling to the LLM-elicited distributions, raising probabilities to a power of $1/\tau$ and re-normalizing: $\hat{p}_\tau(y=k) \propto \hat{p}(y=k)^{1/\tau}$, where $\tau = 1$ recovers the original distribution, $\tau < 1$ sharpens it, and $\tau > 1$ softens it. Results for $\tau \in \{0.5, 2\}$ show that both sharpening and softening consistently degrade performance. Sharpening ($\tau = 0.5$) can initially appear competitive at very small budgets for the nonadaptive method, but degrades sharply as budget increases due to overconfident likelihoods that amplify posterior misassignment. Softening ($\tau = 2$) uniformly underperforms by washing out the discriminative signal in the response distributions. These results suggest that the original LLM-elicited distributions are already well-calibrated for the adaptive querying objective, and that post-hoc rescaling is unlikely to improve performance without additional task-specific calibration data.

\begin{table}[ht]
\caption{Ablation study: log loss for \textbf{nonadaptive} design on real users ($N = 10{,}000$ sampled); cells report mean with standard error below. \textbf{Bold} marks the best value per column.}
\label{tab:ablation_nonadaptive}
\centering
\small
\begin{tabular}{l*{7}{c}}
\toprule
\textbf{Variant} & $T\!=\!5$ & $T\!=\!10$ & $T\!=\!15$ & $T\!=\!20$ & $T\!=\!30$ & $T\!=\!50$ & $T\!=\!86$ \\
\midrule
current setup     & \cell{1.022}{.006} & \cell{1.000}{.006} & \cell{\textbf{.976}}{.006} & \cell{\textbf{.962}}{.006} & \cell{\textbf{.961}}{.006} & \cell{\textbf{.968}}{.006} & \cell{.979}{.006} \\[4pt]
cluster = 50      & \cell{1.023}{.006} & \cell{1.005}{.006} & \cell{.986}{.006} & \cell{.977}{.006} & \cell{.970}{.006} & \cell{.970}{.006} & \cell{.977}{.006} \\[4pt]
cluster = 200     & \cell{1.022}{.006} & \cell{.997}{.006} & \cell{.977}{.006} & \cell{.963}{.006} & \cell{.962}{.006} & \cell{\textbf{.968}}{.006} & \cell{\textbf{.976}}{.006} \\[4pt]
\midrule
det.\ $\varepsilon = 0.1$ & \cell{1.029}{.010} & \cell{1.199}{.012} & \cell{1.272}{.013} & \cell{1.291}{.014} & \cell{1.351}{.014} & \cell{1.421}{.015} & \cell{1.472}{.015} \\[4pt]
det.\ $\varepsilon = 0.3$ & \cell{1.080}{.006} & \cell{1.064}{.007} & \cell{1.081}{.007} & \cell{1.106}{.008} & \cell{1.116}{.008} & \cell{1.136}{.008} & \cell{1.168}{.009} \\[4pt]
\midrule
temp $\tau = 0.5$ & \cell{\textbf{1.005}}{.009} & \cell{\textbf{.981}}{.009} & \cell{1.022}{.010} & \cell{1.021}{.010} & \cell{1.034}{.011} & \cell{1.071}{.011} & \cell{1.085}{.011} \\[4pt]
temp $\tau = 2$   & \cell{1.097}{.004} & \cell{1.087}{.004} & \cell{1.075}{.004} & \cell{1.074}{.004} & \cell{1.071}{.004} & \cell{1.068}{.004} & \cell{1.069}{.004} \\
\bottomrule
\end{tabular}
\end{table}

\begin{table}[ht]
\caption{Ablation study: log loss for \textbf{greedy} design on real users ($N = 10{,}000$ sampled); cells report mean with standard error below. \textbf{Bold} marks the best value per column.}
\label{tab:ablation_greedy}
\centering
\small
\begin{tabular}{l*{7}{c}}
\toprule
\textbf{Variant} & $T\!=\!5$ & $T\!=\!10$ & $T\!=\!15$ & $T\!=\!20$ & $T\!=\!30$ & $T\!=\!50$ & $T\!=\!86$ \\
\midrule
current setup     & \cell{\textbf{1.003}}{.006} & \cell{.993}{.006} & \cell{.987}{.006} & \cell{\textbf{.984}}{.006} & \cell{\textbf{.984}}{.006} & \cell{.984}{.006} & \cell{.979}{.006} \\[4pt]
cluster = 50      & \cell{1.012}{.006} & \cell{.999}{.006} & \cell{.996}{.006} & \cell{.990}{.006} & \cell{.985}{.006} & \cell{.981}{.006} & \cell{.977}{.006} \\[4pt]
cluster = 200     & \cell{1.004}{.006} & \cell{\textbf{.992}}{.006} & \cell{\textbf{.986}}{.006} & \cell{.986}{.006} & \cell{.986}{.006} & \cell{\textbf{.978}}{.006} & \cell{\textbf{.976}}{.006} \\[4pt]
\midrule
det.\ $\varepsilon = 0.1$ & \cell{1.218}{.012} & \cell{1.329}{.013} & \cell{1.368}{.014} & \cell{1.399}{.014} & \cell{1.427}{.015} & \cell{1.490}{.015} & \cell{1.472}{.015} \\[4pt]
det.\ $\varepsilon = 0.3$ & \cell{1.065}{.007} & \cell{1.085}{.007} & \cell{1.106}{.008} & \cell{1.121}{.008} & \cell{1.138}{.008} & \cell{1.167}{.009} & \cell{1.168}{.009} \\[4pt]
\midrule
temp $\tau = 0.5$ & \cell{1.052}{.010} & \cell{1.078}{.011} & \cell{1.097}{.011} & \cell{1.094}{.012} & \cell{1.107}{.012} & \cell{1.101}{.012} & \cell{1.085}{.011} \\[4pt]
temp $\tau = 2$   & \cell{1.098}{.004} & \cell{1.088}{.004} & \cell{1.082}{.004} & \cell{1.078}{.004} & \cell{1.075}{.004} & \cell{1.072}{.004} & \cell{1.069}{.004} \\
\bottomrule
\end{tabular}
\end{table}

\subsection{Runtime Comparison}

Table~\ref{tab:runtime} reports wall-clock runtimes for all methods on the real WorldValuesBench dataset (70{,}767 training users, 17{,}692 test users, $T = 86$). All implementations are optimized with standard techniques including contiguous NumPy arrays, Numba JIT compilation, and Joblib parallelization, and all timings were measured on a single Apple MacBook Pro (M1 chip, 8-core CPU/GPU, 16\,GB memory). The table separates \emph{inference} (online computation on test users) from \emph{fitting} (offline model calibration on training users). All persona-based methods share a single empirical Bayes prior fitting step, described in the prior-specification paragraph above, which runs an EM algorithm with a maximum of 100 iterations and convergence tolerance $10^{-4}$, completing in 3.98 minutes. Because this fitting is performed only once, the total cost of running multiple persona-based strategies is the sum of inference times across methods plus a single 3.98-minute fitting cost. In contrast, each CAT baseline requires its own item parameter calibration, so fitting costs cannot be shared across CAT variants; implementation details for CAT methods are in Appendix~\ref{app:cat:implementation}.

\begin{table}[ht]
\caption{Runtime comparison on real WorldValuesBench ($n_{\mathrm{train}} = 70{,}767$, $n_{\mathrm{test}} = 17{,}692$; $T = 86$). Inference = online computation on test users; Fitting = offline calibration on training users. $^\dagger$Persona-based methods share a single fitting step; the Total column reports the cost of running each method individually.}
\label{tab:runtime}
\centering
\small
\begin{tabular}{lccc}
\toprule
\textbf{Method} & \textbf{Inference (min)} & \textbf{Fitting (min)} & \textbf{Total (min)} \\
\midrule
full          & 0.46 & 3.98$^\dagger$ & 4.44 \\
random\_fixed & 0.47 & 3.98$^\dagger$ & 4.45 \\
nonadaptive   & 0.50 & 3.98$^\dagger$ & 4.48 \\
random        & 0.50 & 3.98$^\dagger$ & 4.48 \\
greedy        & 40.36 & 3.98$^\dagger$ & 44.34 \\
\midrule
CAT-GRM       & 10.05 & 10.64 & 20.69 \\
CAT-GPCM      & 7.85  & 14.33 & 22.18 \\
CAT-MGRM ($D\!=\!3$) & 27.52 & 67.82 & 95.34 \\
CAT-MGPCM ($D\!=\!3$) & 34.61 & 124.42 & 159.03 \\
\bottomrule
\end{tabular}
\end{table}

Non-adaptive persona-based methods (\textbf{full}, \textbf{random\_fixed}, \textbf{nonadaptive}, \textbf{random}) complete inference for all 17{,}692 test users in under one minute. Including the shared fitting cost of 3.98 minutes, the total wall-clock time for any single non-adaptive persona method is under five minutes. The \textbf{greedy} method requires ${\sim}40$ minutes for inference because it recomputes the one-step lookahead objective at every step for every user, bringing its total to ${\sim}44$ minutes---still practical for moderate-scale applications. Crucially, because all persona-based methods reuse the same fitted prior, the 3.98-minute fitting cost is incurred only once rather than once per strategy.

Among the CAT baselines, the unidimensional models (GRM, GPCM) require ${\sim}8$--$10$ minutes for inference and ${\sim}10$--$15$ minutes for item parameter fitting, totaling ${\sim}20$--$22$ minutes each. The multidimensional models (MGRM, MGPCM) are substantially more expensive: MGRM totals ${\sim}95$ minutes and MGPCM ${\sim}159$ minutes, with the majority of the cost attributable to offline fitting on a $D=3$-dimensional Cartesian grid. Unlike persona-based methods, each CAT variant requires its own item parameter calibration, so running all four CAT baselines costs ${\sim}297$ minutes. Notably, these multidimensional CAT baselines use only $D=3$ latent dimensions, yet already incur $4$--$7{\times}$ the total runtime of unidimensional CAT. Scaling MIRT to higher dimensions is computationally prohibitive due to the exponential growth of the grid. This illustrates a fundamental expressiveness--scalability tradeoff in classical CAT: while multidimensional IRT captures richer latent structure and does yield improved predictions over unidimensional models (Tables~\ref{tab:synthetic_logloss_main} and~\ref{tab:real_logloss}), the computational cost of increasing $D$ grows rapidly and limits the practical dimensionality of the latent space. In contrast, the persona-based model achieves its expressiveness through a large dictionary of $n=2{,}058$ semantically grounded LLM-powered personas, while maintaining the same lightweight closed-form inference.

\section{Related Work}\label{sec:related}

\paragraph{Bayesian experimental design (BED).}
BED dates back to the seminal work of \citet{lindley1956measure} and has since developed into a rich and mature literature \citep{chaloner1995bayesian, rainforth2024modern}. The central idea is to select experiments or queries that optimize an information-theoretic or decision-theoretic objective under a Bayesian model. While conceptually powerful, classical BED methods are often computationally demanding, typically requiring nested Monte Carlo estimation or variational approximations of posterior quantities. As a result, even approximate implementations can be expensive at scale, and exact posterior inference is rarely tractable in high-dimensional settings. Recent work has pursued neural and amortized variants of sequential BED---including mutual-information neural estimation \citep{kleinegesse2020bayesian} and learned design policies for real-time deployment \citep{foster2021deep, ivanova2021implicit}---but these approaches replace exact inference with learned surrogates or policy networks. In contrast, our persona-induced mixture model retains closed-form posterior updates while leveraging the expressiveness of LLM-generated response distributions. Moreover, the classical BED literature has traditionally relied on parametric statistical models and does not leverage modern generative models as components of the prior or likelihood.

\paragraph{Active learning and noisy Bayesian querying.}
Active learning has a long history, with a comprehensive overview provided by \citet{settles2009active}. Our problem is most closely related to Bayesian active learning with noisy observations, where the learner adaptively selects queries to reduce uncertainty about latent structure. Some prior works consider conceptually related problems but differ substantially in formulation or assumptions. For example, \citet{bruno2012twenty} study a setting where queries ask whether an item belongs to a proposed set, which does not directly apply to our multi-question, multi-response framework. The EC2 framework of \citet{golovin2010near} considers adaptive querying over a hypothesis space, but is primarily designed for noiseless responses and a relatively small number of hypotheses, making it unsuitable for our setting with a large pool of personas and inherently noisy responses. More broadly, several works in noisy Bayesian active learning, such as \citet{naghshvar2012noisy}, rely on assumptions that no two hypotheses are indistinguishable forever. In contrast, in our setting, different personas may remain probabilistically indistinguishable even after exhausting the querying budget. Our formulation is also related to best-arm identification in bandit problems, but differs in that our objective is not to identify a single optimal arm with i.i.d. rewards, but rather to minimize a general posterior objective functional that may depend on high-dimensional response distributions or downstream decision quality.

\paragraph{CAT and IRT.}
Computerized adaptive testing (CAT) and item response theory (IRT) provide a well-established framework for adaptively selecting test items to efficiently estimate a test-taker's latent trait \citep{wainer2000computerized, van2010elements, lord2012applications}. Bayesian approaches to CAT can often be viewed as special cases of BED, with objectives such as posterior variance reduction or information maximization. However, classical CAT and item response theory (IRT) models typically rely on low-dimensional latent variables---often a single scalar ability parameter---and parametric item response functions. These modeling assumptions can be restrictive when user characteristics and response patterns are complex or heterogeneous. Furthermore, posterior updates and predictive likelihoods become intractable in higher-dimensional extensions \citep{reckase2009multidimensional} or nonparametric variants, limiting scalability. In practice, CAT methods also require a costly offline calibration phase to fit item parameters from large datasets \citep{bock1981marginal}, which may not transfer well across domains.

\paragraph{Latent class analysis and finite mixtures.}
Our model is also related to latent class analysis \citep{goodman1974exploratory} and finite mixture models \citep{mclachlan2000finite} for multivariate categorical data. The closest structural analogy is a discrete latent class model with class-conditional question-response probabilities. While classical LCA estimates both the class proportions and the class-conditional distributions entirely from respondent data, our approach specifies the class dictionary and class-conditional distributions \emph{offline} using LLM-generated personas. This distinction is critical for cold-start settings where respondent data for a new item is limited or even unavailable. More broadly, our approach connects to a recent line of work on using generative AI to articulate Bayesian priors. \citet{ohagan2025ai} propose taking a generative AI model as the base measure of a Dirichlet process prior on the data-generating distribution, and performing nonparametric loss-based inference via a parallelizable posterior bootstrap. Our setting differs in that LLM persona prompting yields a finite, interpretable mixture prior tailored to multi-question, multi-response elicitation, which admits closed-form posterior updates suitable for adaptive querying.

\paragraph{LLMs as human behavior simulators.}
A growing body of work investigates the use of LLMs to simulate human survey responses and behavioral patterns. Early studies demonstrated that LLMs can replicate aggregate response distributions of demographic subgroups \citep{argyle2023out, aher2023using} and reproduce patterns observed in economic experiments \citep{horton2023large}. Subsequent work has examined which opinions and values are encoded in LLMs \citep{santurkar2023whose, scherrer2023evaluating}, and whether LLMs can serve as reliable proxies for human subjects \citep{gao2025take, hullman2026human}. These investigations reveal both promise and systematic pitfalls: LLM-generated persona responses can capture meaningful variation across subpopulations, but distortions arise from training data biases and the gap between text generation and genuine human cognition \citep{li2025llm, peng2026digitaltwinsfunhousemirrors}. Recent efforts have sought to close this gap through fine-tuning on survey data \citep{cho2024llm, cao2025specializing}, mixture-of-personas architectures for population-level simulation \citep{leng2024reduce, bui2025mixture, wang2026prompts}, synthetic control framework for simulation calibration \citep{fan2026syn}, and formal frameworks for quantifying the information content of LLM-simulated respondents relative to real humans \citep{huang2025uncertainty, iyengar2025model}. Our approach contributes to this line of work by showing how LLM-generated persona response distributions can serve not merely as simulation outputs but as components of an explicit Bayesian prior that supports principled inference and adaptive decision-making.

\paragraph{LLMs for adaptive querying.}
Complementary to their use as simulators, LLMs have also been explored as adaptive natural-language elicitation systems \citep{piriyakulkij2023active, handa2024bayesian, hu2024uncertainty, mazzaccara2024learning, kobalczyk2025active}. These methods typically assume a finite hypothesis set with deterministic or nearly deterministic likelihoods, which in our framework would correspond to a noiseless setting with a small number of personas. In contrast, our setting features inherently stochastic responses where even the true persona's response distribution assigns non-trivial probability to multiple categories, making posterior concentration fundamentally slower and principled uncertainty quantification essential. \citet{wang2025adaptive} propose an adaptive elicitation framework using a meta-learned predictive language model to select questions that maximize simulated future information gain. Our approach differs by maintaining an explicit finite latent persona prior with closed-form Bayesian posterior updates, rather than relying on predictive uncertainty from a neural sequence model. 

\section{Discussion}\label{sec:discussion}

Our results suggest that AI personas offer a practical middle ground between classical parametric latent variable models and fully black-box generative approaches. By encoding rich prior knowledge through LLM-simulated persona--question response distributions, the resulting model remains expressive while admitting closed-form Bayesian updates and efficient predictive inference. This tractability enables the direct use of standard Bayesian experimental design and adaptive querying methods without resorting to expensive posterior approximations.

\paragraph{Comparison with CAT and cold-start regimes.}
While CAT provides a natural baseline, its reliance on low-dimensional latent traits and offline calibration limits its effectiveness in cold-start user or item settings. In contrast, persona-based priors inject structured prior information that can be leveraged immediately, even when personas are not derived from the evaluation dataset. The observed gains therefore arise not from fitting a more complex model, but from better prior specification within a Bayesian framework.

\paragraph{Calibration of LLM-elicited distributions.}
A natural concern is whether directly elicited LLM probability distributions are well-calibrated. Our ablation studies provide indirect but encouraging evidence: replacing the elicited distributions with deterministic responses plus uniform noise consistently degrades performance across all methods and metrics, as does applying temperature scaling to reshape the distributions. This suggests that the distributional shape of the LLM-elicited responses carries useful information beyond the modal answer, even if the distributions are not perfectly calibrated in an absolute sense. Developing principled calibration procedures for persona-conditioned response distributions---for example, using a small amount of validation data to learn a calibration map---remains an important direction for future work.

\paragraph{Limitations.}
Our approach depends on the quality and diversity of the persona dictionary and the fidelity of LLM-generated response distributions, and we currently assume categorical responses and conditional independence given the persona. Furthermore, LLM-generated persona priors may encode or amplify biases present in the LLM's training data. If the persona dictionary underrepresents certain demographic or attitudinal groups, the resulting prior will systematically assign low probability to those users, potentially leading to poor predictions and inequitable outcomes. Even when the dictionary is diverse, the LLM-elicited response distributions may reflect stereotypical rather than genuine response patterns for underrepresented groups. Developing calibration and debiasing procedures for persona priors---including auditing persona dictionaries for representational balance and validating elicited distributions against ground-truth subgroup data---is an important direction for responsible deployment.

\paragraph{Future directions.}
More broadly, extending the framework to richer response types, learned persona dictionaries, or explicit dependencies across questions remains an important direction for future work. Several extensions are particularly promising: (i) combining persona priors with parametric models, for example using persona posteriors as warm starts for IRT, could leverage the strengths of both approaches; (ii) learning or refining the persona dictionary over time from observed user data would enable the model to adapt to new populations; and (iii) applications beyond surveys---including recommender systems, medical questionnaires, and intelligent tutoring systems---represent natural domains where the cold-start advantages of persona priors could prove valuable.

\section*{Acknowledgements}
Kaizheng Wang's research is supported by NSF grant DMS-2515679.

\bibliography{bib}

@article{mccarthy1956measures,
author = {John McCarthy},
title = {Measures of the Value of Information},
journal = {Proceedings of the National Academy of Sciences},
volume = {42},
number = {9},
pages = {654-655},
year = {1956},
}

@article{savage1971elicitation,
author = {Leonard J. Savage},
title = {Elicitation of Personal Probabilities and Expectations},
journal = {Journal of the American Statistical Association},
volume = {66},
number = {336},
pages = {783--801},
year = {1971},
}

@article{gneiting2007strictly,
author = {Tilmann Gneiting and Adrian E Raftery},
title = {Strictly Proper Scoring Rules, Prediction, and Estimation},
journal = {Journal of the American Statistical Association},
volume = {102},
number = {477},
pages = {359--378},
year = {2007},
}

@inproceedings{rainforth2018nesting,
  title     = {On Nesting Monte Carlo Estimators},
  author    = {Rainforth, Tom and Cornish, Rob and Yang, Hongseok and Warrington, Andrew and Wood, Frank},
  booktitle = {Proceedings of the 35th International Conference on Machine Learning},
  pages     = {4267--4276},
  year      = {2018},
  publisher = {PMLR},
  series    = {Proceedings of Machine Learning Research},
}

@inproceedings{foster2019variational,
  title     = {Variational Bayesian Optimal Experimental Design},
  author    = {Foster, Adam and Jankowiak, Martin and Bingham, Eli and Horsfall, Paul and Teh, Yee Whye and Rainforth, Tom and Goodman, Noah},
  booktitle = {Advances in Neural Information Processing Systems},
  volume    = {32},
  year      = {2019},
}

@misc{zhao2024worldvaluesbench,
      title={WorldValuesBench: A Large-Scale Benchmark Dataset for Multi-Cultural Value Awareness of Language Models}, 
      author={Wenlong Zhao and Debanjan Mondal and Niket Tandon and Danica Dillion and Kurt Gray and Yuling Gu},
      year={2024},
      eprint={2404.16308},
      archivePrefix={arXiv},
      primaryClass={cs.CL},
}

@inproceedings{golovin2010near,
author = {Golovin, Daniel and Krause, Andreas and Ray, Debajyoti},
title = {Near-optimal Bayesian active learning with noisy observations},
year = {2010},
booktitle = {Proceedings of the 24th International Conference on Neural Information Processing Systems - Volume 1},
pages = {766-774},
numpages = {9},
series = {NIPS'10}
}

@article{lindley1956measure,
  title = {On a {{Measure}} of the {{Information Provided}} by an {{Experiment}}},
  author = {Lindley, D. V.},
  year = 1956,
  month = dec,
  journal = {The Annals of Mathematical Statistics},
  volume = {27},
  number = {4},
  pages = {986--1005},
  issn = {0003-4851},
}

@article{chaloner1995bayesian,
author = {Kathryn Chaloner and Isabella Verdinelli},
title = {{Bayesian Experimental Design: A Review}},
volume = {10},
journal = {Statistical Science},
number = {3},
publisher = {Institute of Mathematical Statistics},
pages = {273 -- 304},
year = {1995},
}

@article{rainforth2024modern,
author = {Tom Rainforth and Adam Foster and Desi R. Ivanova and Freddie Bickford Smith},
title = {{Modern Bayesian Experimental Design}},
volume = {39},
journal = {Statistical Science},
number = {1},
publisher = {Institute of Mathematical Statistics},
pages = {100 -- 114},
year = {2024},
}

@techreport{settles2009active,
  author = {Settles, Burr},
  title  = {Active Learning Literature Survey},
  year   = {2009},
  institution = {University of Wisconsin--Madison}
}

@article{bruno2012twenty,
author = {Bruno Jedynak and Peter I. Frazier and Raphael Sznitman},
title = {{Twenty questions with noise: Bayes optimal policies for entropy loss}},
volume = {49},
journal = {Journal of Applied Probability},
number = {1},
publisher = {Applied Probability Trust},
pages = {114 -- 136},
year = {2012},
}

@inproceedings{naghshvar2012noisy,
  author={Naghshvar, Mohammad and Javidi, Tara and Chaudhuri, Kamalika},
  booktitle={2012 50th Annual Allerton Conference on Communication, Control, and Computing (Allerton)}, 
  title={Noisy Bayesian active learning}, 
  year={2012},
  volume={},
  number={},
  pages={1626-1633},
}

@book{van2010elements,
  title={Elements of adaptive testing},
  author={Van der Linden, Wim J and Glas, Cees AW},
  year={2010},
  publisher={Springer}
}

@misc{handa2024bayesian,
      title={Bayesian Preference Elicitation with Language Models}, 
      author={Kunal Handa and Yarin Gal and Ellie Pavlick and Noah Goodman and Jacob Andreas and Alex Tamkin and Belinda Z. Li},
      year={2024},
      eprint={2403.05534},
      archivePrefix={arXiv},
      primaryClass={cs.CL},
}

@inproceedings{piriyakulkij2023active,
  title        = {Active Preference Inference using Language Models and Probabilistic Reasoning},
  author       = {Piriyakulkij, Wasu Top and Kuleshov, Volodymyr and Ellis, Kevin},
  booktitle    = {Proceedings of the Foundation Models for Decision Making Workshop at NeurIPS 2023},
  year         = {2023},
}

@inproceedings{hu2024uncertainty,
  title        = {Uncertainty of Thoughts: Uncertainty-Aware Planning Enhances Information Seeking in Large Language Models},
  author       = {Hu, Zhiyuan and Liu, Chumin and Feng, Xidong and Zhao, Yilun and See-Kiong Ng and Anh Tuan Luu and Junxian He and Pang Wei Koh and Bryan Hooi},
  booktitle    = {Advances in Neural Information Processing Systems 38},
  year         = {2024},
  pages        = {24181--24215},
}

@article{kobalczyk2025active,
  title        = {Active Task Disambiguation with LLMs},
  author       = {Kobalczyk, Katarzyna and Astorga, Nicolas and Liu, Tennison and van der Schaar, Mihaela},
  journal      = {International Conference on Learning Representations (ICLR) 2025},
  year         = {2025},
}

@inproceedings{mazzaccara2024learning,
    title = "Learning to Ask Informative Questions: Enhancing {LLM}s with Preference Optimization and Expected Information Gain",
    author = "Mazzaccara, Davide  and
      Testoni, Alberto  and
      Bernardi, Raffaella",
    editor = "Al-Onaizan, Yaser  and
      Bansal, Mohit  and
      Chen, Yun-Nung",
    booktitle = "Findings of the Association for Computational Linguistics: EMNLP 2024",
    month = nov,
    year = "2024",
    address = "Miami, Florida, USA",
    publisher = "Association for Computational Linguistics",
    pages = "5064--5074",
}

@article{toubia2025database,
  title={Database report: Twin-2k-500: A data set for building digital twins of over 2,000 people based on their answers to over 500 questions},
  author={Toubia, Olivier and Gui, George Z and Peng, Tianyi and Merlau, Daniel J and Li, Ang and Chen, Haozhe},
  journal={Marketing Science},
  volume={44},
  number={6},
  pages={1446--1455},
  year={2025},
  publisher={INFORMS}
}

@article{wang2025adaptive,
  title={Adaptive elicitation of latent information using natural language},
  author={Wang, Jimmy and Zollo, Thomas and Zemel, Richard and Namkoong, Hongseok},
  journal={arXiv preprint arXiv:2504.04204},
  year={2025}
}

@article{argyle2023out,
  title={Out of one, many: Using language models to simulate human samples},
  author={Argyle, Lisa P and Busby, Ethan C and Fulda, Nancy and Gubler, Joshua R and Rytting, Christopher and Wingate, David},
  journal={Political Analysis},
  volume={31},
  number={3},
  pages={337--351},
  year={2023},
  publisher={Cambridge University Press}
}

@inproceedings{aher2023using,
  title={Using large language models to simulate multiple humans and replicate human subject studies},
  author={Aher, Gati V and Arriaga, Rosa I and Kalai, Adam Tauman},
  booktitle={International Conference on Machine Learning},
  pages={337--371},
  year={2023},
  organization={PMLR}
}

@inproceedings{kleinegesse2020bayesian,
  title={Bayesian experimental design for implicit models by mutual information neural estimation},
  author={Kleinegesse, Steven and Gutmann, Michael U},
  booktitle={International Conference on Machine Learning},
  pages={5316--5326},
  year={2020},
  organization={PMLR}
}

@inproceedings{foster2021deep,
  title={Deep adaptive design: Amortizing sequential Bayesian experimental design},
  author={Foster, Adam and Ivanova, Desi R and Malik, Ilyas and Rainforth, Tom},
  booktitle={International Conference on Machine Learning},
  pages={3384--3395},
  year={2021},
  organization={PMLR}
}

@article{ivanova2021implicit,
  title={Implicit deep adaptive design: Policy-based experimental design without likelihoods},
  author={Ivanova, Desi R and Foster, Adam and Kleinegesse, Steven and Gutmann, Michael U and Rainforth, Thomas},
  journal={Advances in Neural Information Processing Systems},
  volume={34},
  pages={25785--25798},
  year={2021}
}

@book{lord2012applications,
  title={Applications of Item Response Theory to Practical Testing Problems},
  author={Lord, Frederic M},
  year={2012},
  publisher={Routledge}
}

@book{wainer2000computerized,
  title={Computerized Adaptive Testing: A Primer},
  author={Wainer, Howard and Dorans, Neil J and Flaugher, Ronald and Green, Bert F and Mislevy, Robert J},
  year={2000},
  publisher={Routledge}
}

@article{reckase2009multidimensional,
  title={Multidimensional item response theory},
  author={Reckase, Mark D},
  journal={Handbook of Statistics},
  volume={26},
  pages={607--642},
  year={2006},
  publisher={Elsevier}
}

@article{bock1981marginal,
  title={Marginal maximum likelihood estimation of item parameters: Application of an {EM} algorithm},
  author={Bock, R Darrell and Aitkin, Murray},
  journal={Psychometrika},
  volume={46},
  number={4},
  pages={443--459},
  year={1981},
  publisher={Springer-Verlag}
}

@article{goodman1974exploratory,
  title={Exploratory latent structure analysis using both identifiable and unidentifiable models},
  author={Goodman, Leo A},
  journal={Biometrika},
  volume={61},
  number={2},
  pages={215--231},
  year={1974},
  publisher={Oxford University Press}
}

@book{mclachlan2000finite,
  title={Finite Mixture Models},
  author={McLachlan, Geoffrey J and Peel, David},
  year={2000},
  publisher={John Wiley \& Sons}
}

@inproceedings{schein2002methods,
  title={Methods and metrics for cold-start recommendations},
  author={Schein, Andrew I and Popescul, Alexandrin and Ungar, Lyle H and Pennock, David M},
  booktitle={Proceedings of the 25th Annual International ACM SIGIR Conference on Research and Development in Information Retrieval},
  pages={253--260},
  year={2002}
}

@techreport{horton2023large,
  title={Large Language Models as Simulated Economic Agents: What Can We Learn from Homo Silicus?},
  author={Horton, John J},
  year={2023},
  institution={National Bureau of Economic Research}
}

@inproceedings{santurkar2023whose,
  title={Whose Opinions Do Language Models Reflect?},
  author={Santurkar, Shibani and Durmus, Esin and Ladhak, Faisal and Lee, Cinoo and Liang, Percy and Hashimoto, Tatsunori},
  booktitle={International Conference on Machine Learning},
  pages={29971--30004},
  year={2023},
  organization={PMLR}
}

@inproceedings{scherrer2023evaluating,
  title={Evaluating the Moral Beliefs Encoded in LLMs},
  author={Scherrer, Nino and Sh, Claudia and Feder, Amir and Blei, David M.},
  booktitle={Proceedings of the 37th International Conference on Neural Information Processing Systems},
  year={2023},
  publisher={Curran Associates Inc.}
}

@article{gao2025take,
  title={Take Caution in Using LLMs as Human Surrogates},
  author={Gao, Yuan and Lee, Dokyun and Burtch, Gordon and Fazelpour, Sina},
  journal={Proceedings of the National Academy of Sciences},
  volume={122},
  number={24},
  pages={e2501660122},
  year={2025},
  publisher={National Academy of Sciences}
}

@article{li2025llm,
  title={LLM Generated Persona is a Promise with a Catch},
  author={Li, Ang and Chen, Haozhe and Namkoong, Hongseok and Peng, Tianyi},
  journal={arXiv preprint arXiv:2503.16527},
  year={2025}
}

@misc{peng2026digitaltwinsfunhousemirrors,
  title={Digital Twins as Funhouse Mirrors: Five Key Distortions},
  author={Peng, Tianyi and Gui, George and Brucks, Melanie and Merlau, Daniel J. and Fan, Grace Jiarui and Ben Sliman, Malek and Johnson, Eric J. and Althenayyan, Abdullah and Bellezza, Silvia and Donati, Dante and Fong, Hortense and Friedman, Elizabeth and Guevara, Ariana and Hussein, Mohamed and Jerath, Kinshuk and Kogut, Bruce and Kumar, Akshit and Lane, Kristen and Li, Hannah and Morwitz, Vicki and Netzer, Oded and Perkowski, Patryk and Toubia, Olivier},
  year={2026},
  eprint={2509.19088},
  archivePrefix={arXiv},
  primaryClass={cs.CY}
}

@article{hullman2026human,
  title={This Human Study Did Not Involve Human Subjects: Validating LLM Simulations as Behavioral Evidence},
  author={Hullman, Jessica and Broska, David and Sun, Huaman and Shaw, Aaron},
  journal={arXiv preprint arXiv:2602.15785},
  year={2026}
}

@inproceedings{cao2025specializing,
  title={Specializing Large Language Models to Simulate Survey Response Distributions for Global Populations},
  author={Cao, Yong and Liu, Haijiang and Arora, Arnav and Augenstein, Isabelle and R{\"o}ttger, Paul and Hershcovich, Daniel},
  booktitle={Proceedings of the 2025 Conference of the Nations of the Americas Chapter of the Association for Computational Linguistics: Human Language Technologies (Volume 1: Long Papers)},
  pages={3141--3154},
  year={2025},
  publisher={Association for Computational Linguistics}
}

@article{cho2024llm,
  author={Cho, Suhyun and Kim, Jaeyun and Kim, Jang Hyun},
  journal={IEEE Access},
  title={LLM-Based Doppelg\"anger Models: Leveraging Synthetic Data for Human-Like Responses in Survey Simulations},
  year={2024},
  volume={12},
  pages={178917-178927}
}

@article{bui2025mixture,
  title={Mixture-of-Personas Language Models for Population Simulation},
  author={Ngoc Bui and Hieu Trung Nguyen and Shantanu Kumar and Julian Theodore and Weikang Qiu and Viet Anh Nguyen and Rex Ying},
  journal={arXiv preprint arXiv:2504.05019},
  year={2025}
}

@article{huang2025uncertainty,
  title={How Many Human Survey Respondents is a Large Language Model Worth? An Uncertainty Quantification Perspective},
  author={Huang, Chengpiao and Wu, Yuhang and Wang, Kaizheng},
  journal={arXiv preprint arXiv:2502.17773},
  year={2025}
}

@article{leng2024reduce,
  title={Reduce Disparity Between LLMs and Humans: Optimal LLM Sample Calibration},
  author={Leng, Yan and Sang, Yunxin and Agarwal, Ashish},
  journal={SSRN Working Paper 4802019},
  year={2024}
}

@article{wang2026prompts,
  title={Prompts to Proxies: Emulating Human Preferences via a Compact LLM Ensemble},
  author={Wang, Bingchen and Khoo, Zi-Yu and Wang, Jingtan},
  journal={arXiv preprint arXiv:2509.11311},
  year={2026}
}

@article{fan2026syn,
  title={SYN-DIGITS: A Synthetic Control Framework for Calibrated Digital Twin Simulation},
  author={Fan, Grace Jiarui and Huang, Chengpiao and Peng, Tianyi and Wang, Kaizheng and Wu, Yuhang},
  journal={arXiv preprint arXiv:2604.07513},
  year={2026}
}

@article{samejima1969estimation,
  title={Estimation of latent ability using a response pattern of graded scores},
  author={Samejima, Fumiko},
  journal={Psychometrika},
  volume={34},
  number={S1},
  pages={1--97},
  year={1969},
  publisher={Springer-Verlag}
}

@article{muraki1992generalized,
  title={A generalized partial credit model: Application of an EM algorithm},
  author={Muraki, Eiji},
  journal={ETS Research Report Series},
  volume={1992},
  number={1},
  pages={i--30},
  year={1992},
  publisher={Wiley Online Library}
}

@article{yao2006multidimensional,
  title={A multidimensional partial credit model with associated item and test statistics: An application to mixed-format tests},
  author={Yao, Lihua and Schwarz, Richard D},
  journal={Applied psychological measurement},
  volume={30},
  number={6},
  pages={469--492},
  year={2006},
  publisher={Sage Publications Sage CA: Thousand Oaks, CA}
}

@article{su2009survey,
  title={A survey of collaborative filtering techniques},
  author={Su, Xiaoyuan and Khoshgoftaar, Taghi M},
  journal={Advances in artificial intelligence},
  volume={2009},
  number={1},
  pages={421425},
  year={2009},
  publisher={Wiley Online Library}
}

@article{ohagan2025ai,
  title={AI-Powered Bayesian Inference},
  author={O'Hagan, Sean and Ro{\v{c}}kov{\'a}, Veronika},
  journal={arXiv preprint arXiv:2502.19231},
  year={2025}
}

@article{iyengar2025model,
  title={Model-Free Assessment of Simulator Fidelity via Quantile Curves},
  author={Iyengar, Garud and Lin, Yu-Shiou Willy and Wang, Kaizheng},
  journal={arXiv preprint arXiv:2512.05024},
  year={2025}
}
\bibliographystyle{ims}

\appendix

\section{Notation}\label{app:notation}

Table~\ref{tab:notation} summarizes key notation used throughout the paper.

\begin{table}[ht]
\caption{Summary of notation.}
\label{tab:notation}
\centering
\small
\begin{tabular}{ll}
\toprule
\textbf{Symbol} & \textbf{Description} \\
\midrule
$m$ & Number of questions in the bank \\
$K$ & Number of response categories per question \\
$\mathcal{Y} = \{1,\dots,K\}$ & Response space \\
$\Y \in \mathcal{Y}^m$ & Random response vector \\
$Y_x$ & Response to question $x$ \\
$\mathcal{I}_{\mathrm{feas}}$ & Feasible question set \\
$I^\star$ & Target question set \\
$T$ & Query budget \\
$h_t$ & Interaction history at step $t$ \\
$P_t = p(Z \mid h_t)$ & Posterior distribution at step $t$ \\
$U(\cdot)$ & Uncertainty functional \\
$S(p,z)$ & Scoring rule \\
\midrule
$n$ & Number of personas in dictionary \\
$\theta \in \{1,\dots,n\}$ & Persona membership (latent variable) \\
$\xi_\theta$ & Textual profile of persona $\theta$ \\
$\mu_{\theta,x} \in \Delta^{K-1}$ & Persona--question response distribution \\
$p(\theta)$ & Prior over persona membership \\
$p(\theta \mid \Y_{I_t})$ & Posterior over persona membership \\
$\Delta_U(x \mid h_t)$ & Expected posterior uncertainty of querying $x$ \\
\bottomrule
\end{tabular}
\end{table}

\section{Obtaining Response Distributions from LLMs}\label{app:response_dist}

The persona-induced latent variable model requires, for each persona $\xi_\theta$ and question $x$, a probabilistic response model
$
\mathsf{LLM}(\xi_\theta, x),
$
i.e., a distribution over the possible answers to $x$ when conditioned on persona $\xi_\theta$. While LLMs are typically accessed as conditional text generators, there are multiple ways to obtain or approximate such response distributions. We briefly summarize several common strategies and discuss their trade-offs.

\begin{itemize}
    \item \textbf{Direct distribution elicitation.}
    One approach is to directly prompt the LLM to output a probability distribution over the admissible responses (e.g., normalized probabilities over Likert categories). This method is simple and inexpensive, and works well when the response space is small and well-defined. However, the resulting distributions may be poorly calibrated or sensitive to prompt phrasing, and there is limited theoretical grounding for treating the reported probabilities as true likelihoods.

    \item \textbf{Logit-based extraction.}
    When available, one can extract next-token logits corresponding to each admissible response and normalize them to form a distribution. This approach provides a more direct connection to the underlying language model and avoids heuristic prompting. However, access to token-level logits is restricted or unavailable for many state-of-the-art models, and mapping natural-language responses to token probabilities can be nontrivial.

    \item \textbf{Repeated sampling.}
    Another option is to sample multiple responses from the LLM under a fixed prompt and estimate an empirical distribution over answers. Because persona--question pairs are independent, this procedure can be performed offline and parallelized. Nonetheless, achieving low-variance estimates may require a large number of samples, making this approach computationally expensive at scale.

    \item \textbf{Deterministic response with injected noise.}
    A simpler alternative is to take a deterministic (e.g., temperature-$0$) response and inject synthetic noise to form a distribution. While computationally cheap, this method often produces unrealistic or overly concentrated distributions, particularly when the response space is multi-modal or when subtle preference uncertainty matters.
\end{itemize}
In our experiments, we adopt the direct distribution elicitation approach, as it provides a practical trade-off between computational cost and expressiveness for small categorical response spaces. We also perform ablations that employ the deterministic-plus-noise strategy. We emphasize, however, that our framework is agnostic to the specific method used to obtain $\mathsf{LLM}(\xi_\theta, x)$, and any approach that yields a valid conditional distribution can be plugged into the model.

\section{Prompting Details for LLM Response Distributions}\label{app:llm_prompts}
To obtain the response distributions $\mathsf{LLM}(\xi_\theta, x)$ for each persona-question pair, we use the following prompt template when querying GPT-5-mini. Formats are altered for better readability.

\begin{tcolorbox}[prompttemplate, breakable, title=System Prompt]
You are an expert in simulating human survey responses. You will be given:
\begin{itemize}
\item a detailed persona profile describing a human's values, beliefs, and background;
\item a survey question with ordinal response options numbered 1 to 4.
\end{itemize}
Your task is to predict the persona's *response distribution* to the question.

\vspace{0.5em}
Important instructions:
\begin{itemize}
\item Responses are **ordinal**: higher numbers indicate stronger agreement, endorsement, or intensity (as implied by the question).
\item Output a probability distribution over responses \{1,2,3,4\}.
\item The distribution should reflect realistic human uncertainty: do NOT assume the persona always responds deterministically.
\item If the persona strongly aligns with one side, assign higher probability there, but still allow nonzero probability for nearby options.
\item The probabilities must be non-negative and sum to exactly 1.
\item Avoid assigning probability 1.0 or 0.0 unless the persona makes all other responses essentially impossible.
\end{itemize}
Output format:
Return ONLY a JSON-style list of four numbers: [p1, p2, p3, p4]. Do not include any explanation or additional text.
\end{tcolorbox}

\begin{tcolorbox}[prompttemplate, breakable, title=User Prompt]
PERSONA PROFILE:
\{persona\}

SURVEY QUESTION:
\{question\}

FORMAT INSTRUCTIONS: 
Return ONLY a JSON-style list of four numbers: [p1, p2, p3, p4]. Do not include any explanation or additional text.
\end{tcolorbox}

\section{Additional Results}\label{app:additional_results}

\subsection{Additional Results on WorldValuesBench}\label{app:additional_world}

\begin{figure}[ht]
\centering
\begin{subfigure}{.5\textwidth}
\centering
\includegraphics[width=.95\linewidth]{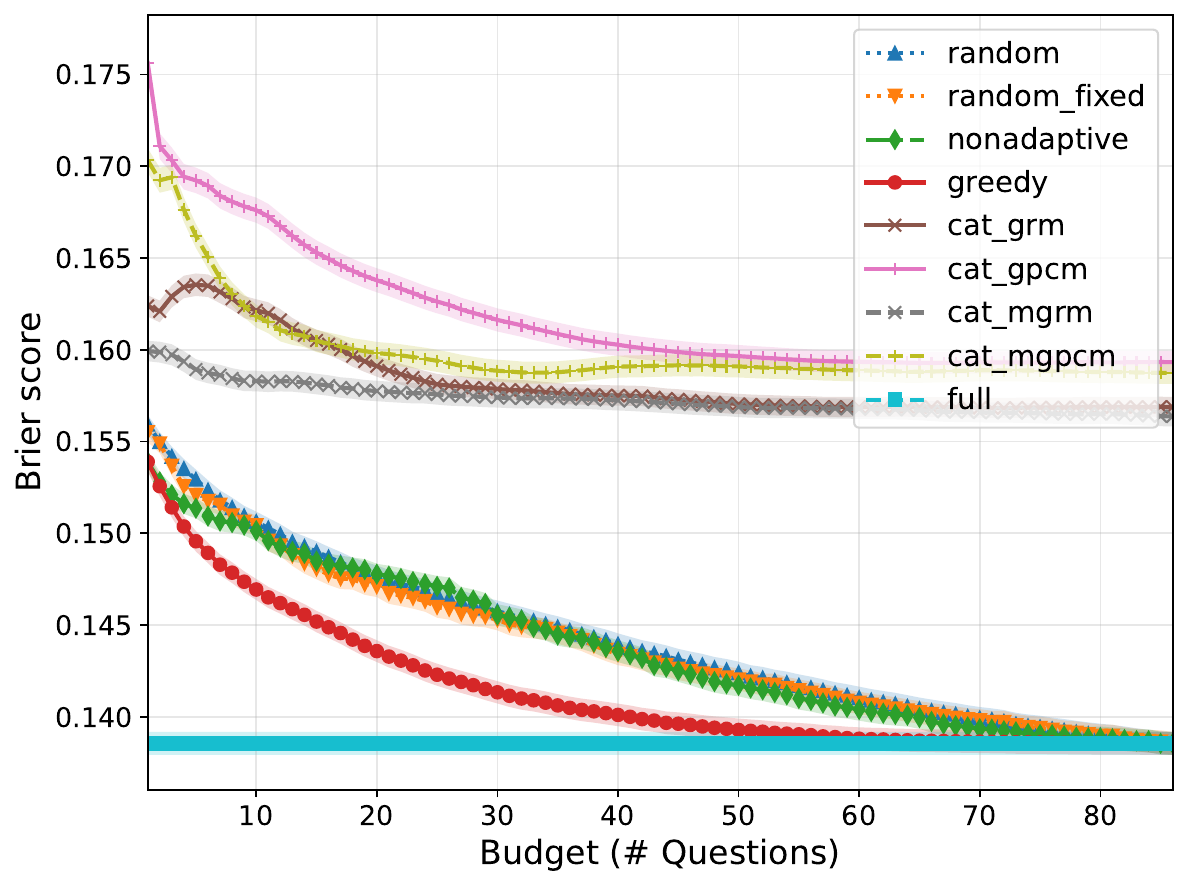}
\caption{Brier score.}
\label{fig:synthetic_results_brier}
\end{subfigure}%
\begin{subfigure}{.5\textwidth}
\centering
\includegraphics[width=.95\linewidth]{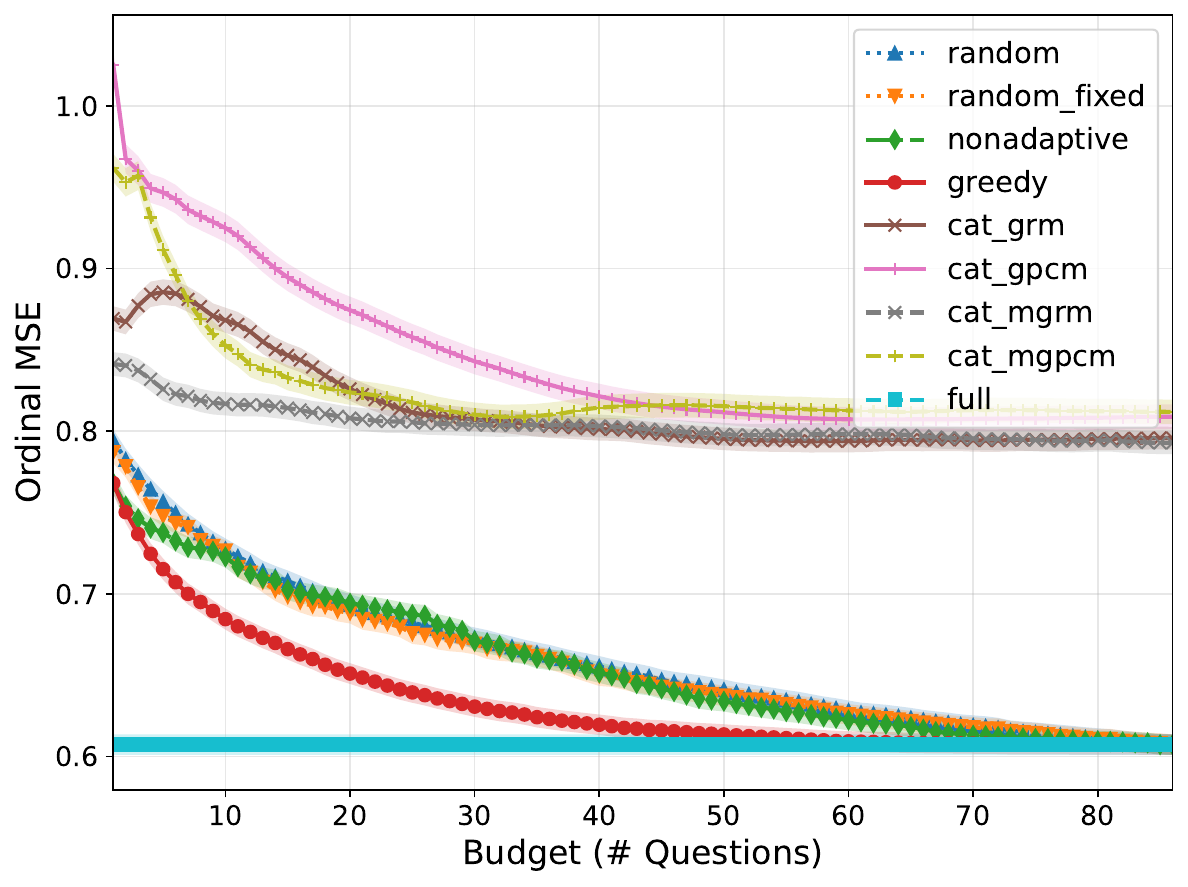}
\caption{Ordinal MSE.}
\label{fig:synthetic_results_mse}
\end{subfigure}
\caption{Synthetic users: performance of all methods as a function of query budget, evaluated using Brier score and ordinal MSE. Persona-based methods substantially outperform CAT baselines, with greedy achieving the strongest performance among persona-based approaches.}
\label{fig:synthetic_results_brier_n_mse}
\end{figure}

\begin{figure}[ht]
\centering
\begin{subfigure}{.5\textwidth}
\centering
\includegraphics[width=.95\linewidth]{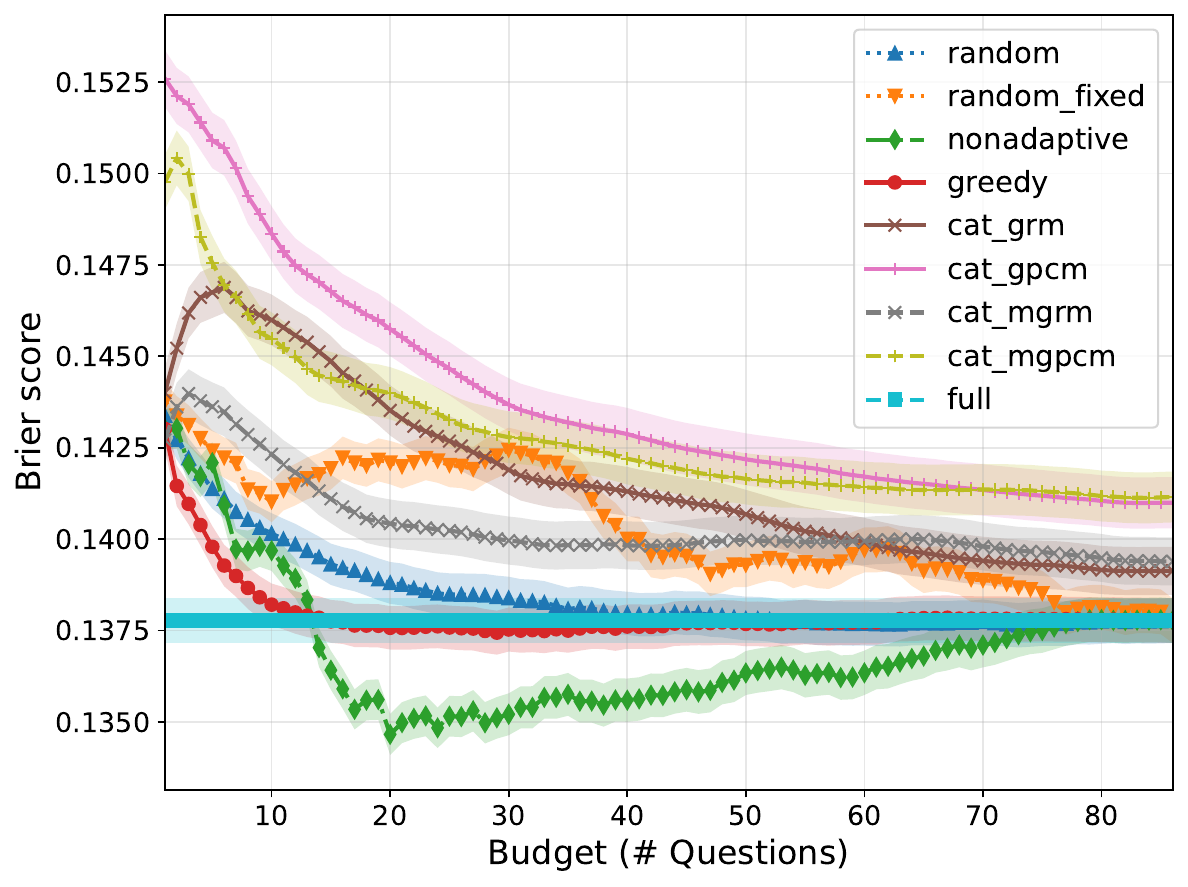}
\caption{Brier score.}
\label{fig:real_results_brier}
\end{subfigure}%
\begin{subfigure}{.5\textwidth}
\centering
\includegraphics[width=.95\linewidth]{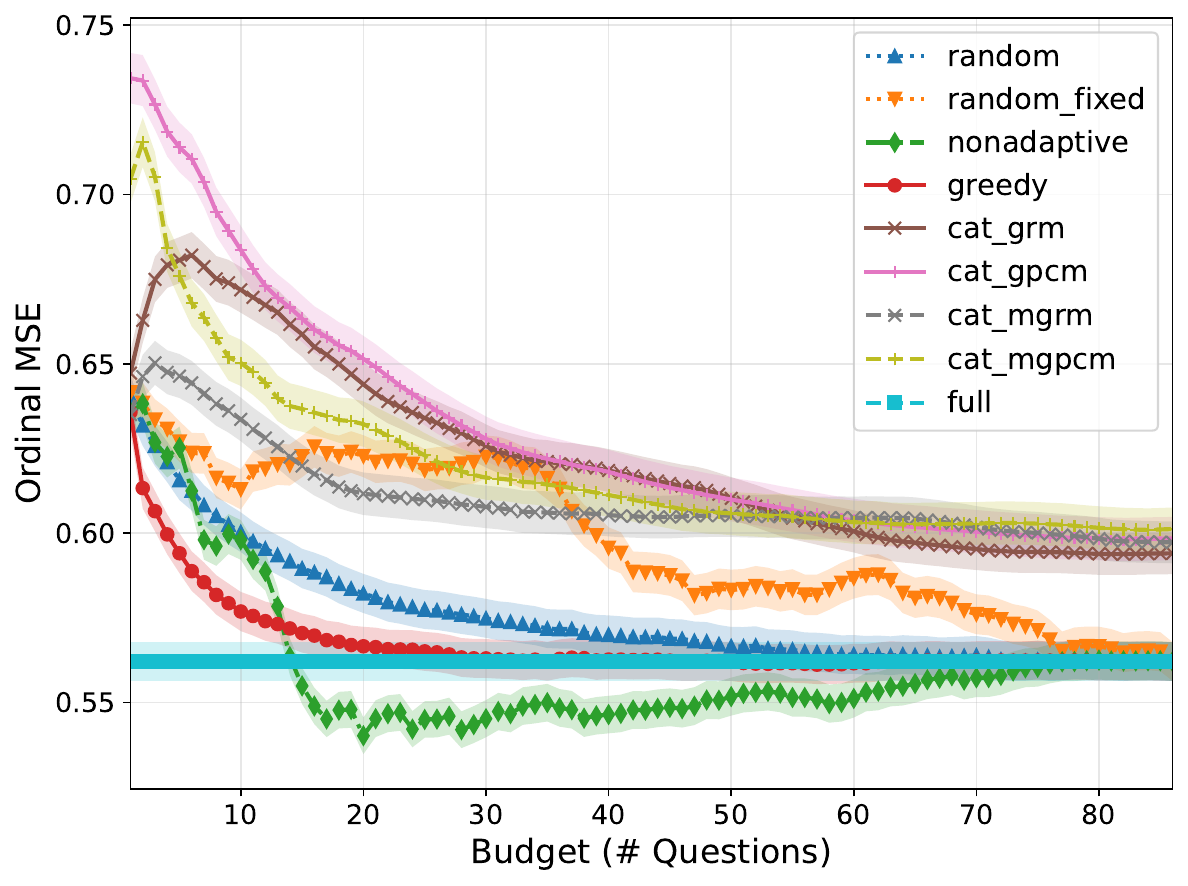}
\caption{Ordinal MSE.}
\label{fig:real_results_mse}
\end{subfigure}
\caption{Real users: performance of all methods as a function of query budget, evaluated using Brier score and ordinal MSE. Persona-based methods outperform CAT baselines; greedy performs best at small budgets, while non-adaptive designs can overtake at larger budgets.}
\label{fig:real_results_brier_n_mse}
\end{figure}

Figures~\ref{fig:synthetic_results_brier_n_mse} and \ref{fig:real_results_brier_n_mse} report additional evaluation results for the held-out question task using Brier score and ordinal MSE, complementing the log-loss results presented in the main text. In all plots, curves denote means averaged over test user--target-question pairs, and shaded regions correspond to 95\% confidence intervals.

Tables~\ref{tab:synthetic_brier} and~\ref{tab:synthetic_mse} report per-budget Brier score and ordinal MSE for synthetic users, complementing the main-text log-loss results (Table~\ref{tab:synthetic_logloss_main}). Tables~\ref{tab:real_brier} and \ref{tab:real_mse} report the per-budget Brier score and ordinal MSE values for real users, complementing the main-text log-loss results (Table~\ref{tab:real_logloss}). 

\begin{table}[ht]
\caption{Synthetic users: Brier score by query budget $T$. $N = 20{,}000$ test users; cells report mean with standard error below. At $T = 86$ all feasible questions have been asked, so all persona-based methods coincide with the \textbf{full} baseline. \textbf{Bold} marks the best value per column.}
\label{tab:synthetic_brier}
\centering
\small
\begin{tabular}{l*{7}{c}}
\toprule
\textbf{Method} & $T\!=\!5$ & $T\!=\!10$ & $T\!=\!15$ & $T\!=\!20$ & $T\!=\!30$ & $T\!=\!50$ & $T\!=\!86$ \\
\midrule
random        & \cell{.1529}{.0003} & \cell{.1506}{.0003} & \cell{.1490}{.0003} & \cell{.1477}{.0003} & \cell{.1458}{.0003} & \cell{.1424}{.0003} & \cell{\textbf{.1386}}{.0003} \\[4pt]
random\_fixed & \cell{.1521}{.0003} & \cell{.1504}{.0003} & \cell{.1481}{.0003} & \cell{.1471}{.0003} & \cell{.1453}{.0003} & \cell{.1420}{.0003} & \cell{\textbf{.1386}}{.0003} \\[4pt]
nonadaptive   & \cell{.1514}{.0003} & \cell{.1501}{.0003} & \cell{.1485}{.0003} & \cell{.1478}{.0003} & \cell{.1456}{.0003} & \cell{.1417}{.0003} & \cell{\textbf{.1386}}{.0003} \\[4pt]
greedy        & \cell{\textbf{.1496}}{.0003} & \cell{\textbf{.1469}}{.0003} & \cell{\textbf{.1452}}{.0003} & \cell{\textbf{.1436}}{.0003} & \cell{\textbf{.1413}}{.0003} & \cell{\textbf{.1393}}{.0003} & \cell{\textbf{.1386}}{.0003} \\[4pt]
\midrule
CAT-GRM       & \cell{.1635}{.0003} & \cell{.1621}{.0003} & \cell{.1605}{.0003} & \cell{.1591}{.0003} & \cell{.1579}{.0003} & \cell{.1570}{.0003} & \cell{.1569}{.0003} \\[4pt]
CAT-GPCM      & \cell{.1692}{.0004} & \cell{.1676}{.0003} & \cell{.1653}{.0003} & \cell{.1638}{.0003} & \cell{.1616}{.0003} & \cell{.1597}{.0003} & \cell{.1593}{.0003} \\[4pt]
CAT-MGRM      & \cell{.1589}{.0003} & \cell{.1583}{.0003} & \cell{.1581}{.0003} & \cell{.1578}{.0003} & \cell{.1574}{.0003} & \cell{.1569}{.0003} & \cell{.1564}{.0003} \\[4pt]
CAT-MGPCM     & \cell{.1662}{.0003} & \cell{.1618}{.0003} & \cell{.1605}{.0003} & \cell{.1598}{.0003} & \cell{.1589}{.0003} & \cell{.1591}{.0003} & \cell{.1587}{.0003} \\
\bottomrule
\end{tabular}
\end{table}

\begin{table}[ht]
\caption{Synthetic users: ordinal MSE by query budget $T$. $N = 20{,}000$ test users; cells report mean with standard error below. At $T = 86$ all feasible questions have been asked, so all persona-based methods coincide with the \textbf{full} baseline. \textbf{Bold} marks the best value per column.}
\label{tab:synthetic_mse}
\centering
\small
\begin{tabular}{l*{7}{c}}
\toprule
\textbf{Method} & $T\!=\!5$ & $T\!=\!10$ & $T\!=\!15$ & $T\!=\!20$ & $T\!=\!30$ & $T\!=\!50$ & $T\!=\!86$ \\
\midrule
random        & \cell{.757}{.003} & \cell{.728}{.003} & \cell{.708}{.003} & \cell{.693}{.003} & \cell{.673}{.003} & \cell{.641}{.003} & \cell{\textbf{.607}}{.003} \\[4pt]
random\_fixed & \cell{.748}{.003} & \cell{.727}{.003} & \cell{.698}{.003} & \cell{.688}{.003} & \cell{.669}{.003} & \cell{.637}{.003} & \cell{\textbf{.607}}{.003} \\[4pt]
nonadaptive   & \cell{.738}{.003} & \cell{.723}{.003} & \cell{.703}{.003} & \cell{.694}{.003} & \cell{.671}{.003} & \cell{.634}{.003} & \cell{\textbf{.607}}{.003} \\[4pt]
greedy        & \cell{\textbf{.715}}{.003} & \cell{\textbf{.684}}{.003} & \cell{\textbf{.666}}{.003} & \cell{\textbf{.651}}{.003} & \cell{\textbf{.631}}{.003} & \cell{\textbf{.614}}{.003} & \cell{\textbf{.607}}{.003} \\[4pt]
\midrule
CAT-GRM       & \cell{.886}{.004} & \cell{.868}{.004} & \cell{.847}{.004} & \cell{.826}{.004} & \cell{.807}{.004} & \cell{.795}{.004} & \cell{.796}{.004} \\[4pt]
CAT-GPCM      & \cell{.947}{.005} & \cell{.925}{.004} & \cell{.895}{.004} & \cell{.874}{.004} & \cell{.843}{.004} & \cell{.812}{.004} & \cell{.809}{.004} \\[4pt]
CAT-MGRM      & \cell{.826}{.004} & \cell{.817}{.004} & \cell{.814}{.004} & \cell{.808}{.004} & \cell{.804}{.004} & \cell{.798}{.004} & \cell{.793}{.004} \\[4pt]
CAT-MGPCM     & \cell{.912}{.004} & \cell{.853}{.004} & \cell{.833}{.004} & \cell{.824}{.004} & \cell{.810}{.004} & \cell{.815}{.004} & \cell{.812}{.004} \\
\bottomrule
\end{tabular}
\end{table}

\begin{table}[ht]
\caption{Real users (WorldValuesBench): Brier score by query budget $T$. $N = 17{,}692$ held-out users; cells report mean with standard error below. At $T = 86$ all feasible questions have been asked, so all persona-based methods coincide with the \textbf{full} baseline. \textbf{Bold} marks the best value per column.}
\label{tab:real_brier}
\centering
\small
\begin{tabular}{l*{7}{c}}
\toprule
\textbf{Method} & $T\!=\!5$ & $T\!=\!10$ & $T\!=\!15$ & $T\!=\!20$ & $T\!=\!30$ & $T\!=\!50$ & $T\!=\!86$ \\
\midrule
random        & \cell{.1414}{.0003} & \cell{.1402}{.0003} & \cell{.1393}{.0003} & \cell{.1388}{.0003} & \cell{.1384}{.0003} & \cell{.1378}{.0003} & \cell{\textbf{.1378}}{.0003} \\[4pt]
random\_fixed & \cell{.1424}{.0003} & \cell{.1410}{.0003} & \cell{.1419}{.0003} & \cell{.1421}{.0003} & \cell{.1424}{.0003} & \cell{.1393}{.0003} & \cell{\textbf{.1378}}{.0003} \\[4pt]
nonadaptive   & \cell{.1421}{.0003} & \cell{.1397}{.0003} & \cell{\textbf{.1364}}{.0003} & \cell{\textbf{.1347}}{.0003} & \cell{\textbf{.1352}}{.0003} & \cell{\textbf{.1363}}{.0003} & \cell{\textbf{.1378}}{.0003} \\[4pt]
greedy        & \cell{\textbf{.1398}}{.0003} & \cell{\textbf{.1382}}{.0003} & \cell{.1377}{.0003} & \cell{.1376}{.0003} & \cell{.1375}{.0003} & \cell{.1377}{.0003} & \cell{\textbf{.1378}}{.0003} \\[4pt]
\midrule
CAT-GRM       & \cell{.1467}{.0004} & \cell{.1460}{.0004} & \cell{.1449}{.0004} & \cell{.1435}{.0004} & \cell{.1419}{.0003} & \cell{.1407}{.0003} & \cell{.1391}{.0003} \\[4pt]
CAT-GPCM      & \cell{.1509}{.0004} & \cell{.1484}{.0004} & \cell{.1468}{.0004} & \cell{.1457}{.0004} & \cell{.1437}{.0004} & \cell{.1422}{.0004} & \cell{.1410}{.0004} \\[4pt]
CAT-MGRM      & \cell{.1436}{.0003} & \cell{.1423}{.0003} & \cell{.1411}{.0003} & \cell{.1404}{.0003} & \cell{.1400}{.0003} & \cell{.1400}{.0003} & \cell{.1394}{.0003} \\[4pt]
CAT-MGPCM     & \cell{.1476}{.0004} & \cell{.1455}{.0004} & \cell{.1444}{.0004} & \cell{.1440}{.0004} & \cell{.1428}{.0004} & \cell{.1416}{.0004} & \cell{.1412}{.0004} \\
\bottomrule
\end{tabular}
\end{table}

\begin{table}[ht]
\caption{Real users (WorldValuesBench): ordinal MSE by query budget $T$. $N = 17{,}692$ held-out users; cells report mean with standard error below. At $T = 86$ all feasible questions have been asked, so all persona-based methods coincide with the \textbf{full} baseline. \textbf{Bold} marks the best value per column.}
\label{tab:real_mse}
\centering
\small
\begin{tabular}{l*{7}{c}}
\toprule
\textbf{Method} & $T\!=\!5$ & $T\!=\!10$ & $T\!=\!15$ & $T\!=\!20$ & $T\!=\!30$ & $T\!=\!50$ & $T\!=\!86$ \\
\midrule
random        & \cell{.616}{.003} & \cell{.600}{.003} & \cell{.590}{.003} & \cell{.582}{.003} & \cell{.575}{.003} & \cell{.567}{.003} & \cell{\textbf{.562}}{.003} \\[4pt]
random\_fixed & \cell{.627}{.003} & \cell{.613}{.003} & \cell{.623}{.003} & \cell{.623}{.003} & \cell{.623}{.003} & \cell{.583}{.003} & \cell{\textbf{.562}}{.003} \\[4pt]
nonadaptive   & \cell{.625}{.003} & \cell{.598}{.003} & \cell{\textbf{.555}}{.003} & \cell{\textbf{.540}}{.003} & \cell{\textbf{.545}}{.003} & \cell{\textbf{.552}}{.003} & \cell{\textbf{.562}}{.003} \\[4pt]
greedy        & \cell{\textbf{.594}}{.003} & \cell{\textbf{.577}}{.003} & \cell{.571}{.003} & \cell{.567}{.003} & \cell{.563}{.003} & \cell{.562}{.003} & \cell{\textbf{.562}}{.003} \\[4pt]
\midrule
CAT-GRM       & \cell{.681}{.003} & \cell{.672}{.003} & \cell{.659}{.003} & \cell{.644}{.003} & \cell{.626}{.003} & \cell{.610}{.003} & \cell{.594}{.003} \\[4pt]
CAT-GPCM      & \cell{.714}{.004} & \cell{.684}{.004} & \cell{.663}{.004} & \cell{.651}{.003} & \cell{.628}{.003} & \cell{.610}{.003} & \cell{.598}{.003} \\[4pt]
CAT-MGRM      & \cell{.646}{.003} & \cell{.634}{.003} & \cell{.620}{.003} & \cell{.612}{.003} & \cell{.608}{.003} & \cell{.605}{.003} & \cell{.597}{.003} \\[4pt]
CAT-MGPCM     & \cell{.676}{.004} & \cell{.650}{.003} & \cell{.637}{.003} & \cell{.632}{.003} & \cell{.616}{.003} & \cell{.606}{.003} & \cell{.601}{.003} \\
\bottomrule
\end{tabular}
\end{table}

\section{CAT Baselines}\label{app:cat_details}

This appendix provides a self-contained overview of computerized adaptive testing (CAT) and the item response theory (IRT) models used as baselines in our experiments. We describe (i) the response models, (ii) parameter estimation via marginal maximum likelihood (MML), (iii) posterior updates during adaptive testing, and (iv) item selection criteria. Throughout, responses take values in $\{0,1,\ldots,K-1\}$.

\subsection{Overview of Computerized Adaptive Testing}

A classical CAT system consists of three components:
\begin{enumerate}
    \item \textbf{Item response model:} a probabilistic model $P(Y_x=k \mid \theta)$ describing how a latent trait $\theta$ (or $\boldsymbol{\theta}$) governs responses to item $x$.
    \item \textbf{Posterior inference:} an update rule for the posterior distribution $p(\theta \mid \mathcal D_t)$ given observed item--response pairs $\mathcal D_t = \{(x_1,y_1),\ldots,(x_t,y_t)\}$.
    \item \textbf{Item selection:} a criterion for selecting the next item $x_{t+1}$ to efficiently reduce uncertainty about $\theta$.
\end{enumerate}
In cognitive testing, $\theta$ typically represents ability. In our survey prediction setting, $\theta$ captures latent attitudes or factors that shape responses. After $T$ adaptive queries, predictions for unasked items are made via the posterior predictive distribution
\[
P(Y_x=k \mid \mathcal D_T)
=
\int P(Y_x=k \mid \theta)\, p(\theta \mid \mathcal D_T)\, d\theta,
\]
or its multidimensional analogue.

\subsection{Item Response Theory Models}

We implement four IRT models: two unidimensional polytomous models (GRM, GPCM) and their multidimensional extensions (MGRM, MGPCM). All models assume conditional independence across items given the latent trait.

\subsubsection{Graded Response Model (GRM)}

The graded response model (GRM) is a cumulative-link model for ordinal responses. For item $x$ with ordered categories $\{0,1,\ldots,K-1\}$, GRM defines cumulative probabilities
\begin{equation}\label{eq:grm_cumulative}
P(Y_x \ge k \mid \theta)
=
\sigma\!\left(a_x(\theta - b_{x,k})\right),
\qquad k=1,\ldots,K-1,
\end{equation}
where $\sigma(u) = 1/(1+e^{-u})$ and $a_x>0$ is the discrimination parameter. The thresholds satisfy
$
b_{x,1} \le b_{x,2} \le \cdots \le b_{x,K-1}.
$
With conventions $P(Y_x \ge 0\mid\theta)=1$ and $P(Y_x \ge K\mid\theta)=0$, category probabilities are obtained by differencing:
\begin{equation}\label{eq:grm_category}
P(Y_x=k \mid \theta)
=
P(Y_x \ge k \mid \theta) - P(Y_x \ge k+1 \mid \theta),
\qquad k=0,\ldots,K-1.
\end{equation}

\subsubsection{Generalized Partial Credit Model (GPCM)}

The generalized partial credit model (GPCM) is an adjacent-category model. One convenient parameterization yields the softmax form
\begin{equation}\label{eq:gpcm_category}
P(Y_x=k \mid \theta)
=
\frac{\exp\!\left(\sum_{s=1}^{k} a_x(\theta - d_{x,s})\right)}
{\sum_{m=0}^{K-1}\exp\!\left(\sum_{s=1}^{m} a_x(\theta - d_{x,s})\right)},
\qquad k=0,\ldots,K-1,
\end{equation}
where $a_x>0$ is discrimination and $\{d_{x,s}\}_{s=1}^{K-1}$ are step parameters (with the convention that an empty sum equals $0$). Unlike GRM, GPCM does not require ordered thresholds.

\subsubsection{Multidimensional GRM (MGRM)}

The multidimensional GRM replaces the scalar trait with $\boldsymbol{\theta}\in\mathbb R^D$ and uses an item discrimination vector $\mathbf a_x\in\mathbb R^D$:
\begin{equation}\label{eq:mgrm_cumulative}
P(Y_x \ge k \mid \boldsymbol{\theta})
=
\sigma\!\left(\mathbf a_x^\top \boldsymbol{\theta} - b_{x,k}\right),
\qquad k=1,\ldots,K-1,
\end{equation}
with category probabilities again computed by differencing consecutive cumulative probabilities.

\subsubsection{Multidimensional GPCM (MGPCM)}

Similarly, the multidimensional GPCM uses $\mathbf a_x\in\mathbb R^D$ and step parameters $\{d_{x,s}\}$:
\begin{equation}\label{eq:mgpcm_category}
P(Y_x=k \mid \boldsymbol{\theta})
=
\frac{\exp\!\left(\sum_{s=1}^{k}\mathbf a_x^\top \boldsymbol{\theta} - d_{x,s}\right)}
{\sum_{m=0}^{K-1}\exp\!\left(\sum_{s=1}^{m}\mathbf a_x^\top \boldsymbol{\theta} - d_{x,s}\right)},
\qquad k=0,\ldots,K-1,
\end{equation}
again using the empty-sum convention. This form reduces to \eqref{eq:gpcm_category} in the unidimensional case.

\subsection{Parameter Estimation via MML (EM)}

IRT parameters are fitted on training users using marginal maximum likelihood (MML). Let $\Y^{(i)}$ denote the responses for user $i$, with missing entries omitted from the product below. Under conditional independence,
\[
p(\Y^{(i)} \mid \theta)
=
\prod_{x \in \mathcal I^{(i)}_{\mathrm{obs}}} P(Y^{(i)}_x \mid \theta),
\]
where $\mathcal I^{(i)}_{\mathrm{obs}}$ is the set of observed items for user $i$. MML maximizes the marginal log-likelihood
\[
\sum_{i=1}^N \log \int p(\Y^{(i)} \mid \theta)\,\phi(\theta)\,d\theta
\]
(or the multidimensional analogue), where $\phi$ denotes the standard normal prior.

\subsubsection{Grid-based approximation}

We approximate integrals over $\theta$ using a fixed grid.
For unidimensional models, we discretize $\theta \in [-\theta_{\max},\theta_{\max}]$ using $G$ grid points $\{\theta^{(g)}\}_{g=1}^G$ with weights $w^{(g)} \propto \phi(\theta^{(g)})$. For multidimensional models with $D$ dimensions, we use a Cartesian grid with $G$ points per dimension, yielding $G^D$ grid points $\{\boldsymbol{\theta}^{(g)}\}_{g=1}^{G^D}$ with weights proportional to the multivariate normal density.

\subsubsection{EM updates}

Given current item parameters, the E-step computes responsibilities
\begin{equation}\label{eq:e_step}
\pi_{i,g}
=
\frac{w^{(g)} \prod_{x \in \mathcal I^{(i)}_{\mathrm{obs}}} P(Y^{(i)}_x \mid \theta^{(g)})}
{\sum_{g'} w^{(g')} \prod_{x \in \mathcal I^{(i)}_{\mathrm{obs}}} P(Y^{(i)}_x \mid \theta^{(g')})}.
\end{equation}
The M-step updates item parameters by maximizing the expected complete-data log-likelihood, separately for each item $x$:
\begin{equation}\label{eq:m_step}
\text{params}(x)
=
\argmax
\sum_{i:\, x \in \mathcal I^{(i)}_{\mathrm{obs}}}
\sum_{g}
\pi_{i,g}\,
\log P(Y^{(i)}_x \mid \theta^{(g)};\text{params}(x)).
\end{equation}
For GRM we enforce $a_x>0$ and ordered thresholds $b_{x,1}\le\cdots\le b_{x,K-1}$; optimization is performed with L-BFGS-B.

\subsection{Posterior Updates During Adaptive Testing}

During adaptive testing, we maintain the posterior over the latent trait on the same grid. Let $w^{(g)}_t$ denote the posterior weight on grid point $\theta^{(g)}$ after observing $\mathcal D_t$.

\paragraph{Initialization.}
We initialize $w^{(g)}_0 \propto \phi(\theta^{(g)})$ (or multivariate normal for MIRT), normalized to sum to $1$.

\paragraph{Bayesian update.}
After querying item $x_t$ and observing response $y_t$, the posterior weights update as
\begin{equation}\label{eq:posterior_update}
w^{(g)}_{t}
=
\frac{w^{(g)}_{t-1}\, P(Y_{x_t}=y_t \mid \theta^{(g)})}
{\sum_{g'} w^{(g')}_{t-1}\, P(Y_{x_t}=y_t \mid \theta^{(g')})}.
\end{equation}

\paragraph{Posterior summaries.}
For 1D models, we compute the posterior mean and variance via
\[
\hat{\theta}_t = \sum_g w_t^{(g)}\theta^{(g)},
\qquad
\mathrm{Var}(\theta \mid \mathcal D_t) = \sum_g w_t^{(g)}(\theta^{(g)}-\hat{\theta}_t)^2,
\]
and similarly compute posterior covariance for multidimensional models.

\subsection{Item Selection Criteria}

We implement standard CAT selection rules. Let $\mathcal I_t$ denote the set of items already administered to the user.

\subsubsection{Maximum Fisher Information (MFI)}

MFI selects the next item by maximizing Fisher information at a point estimate (typically $\hat{\theta}_t$):
\begin{equation}\label{eq:mfi}
x_{t+1}
=
\argmax_{x \in \mathcal I_{\mathrm{feas}}\setminus \mathcal I_t}
I_x(\hat{\theta}_t).
\end{equation}
For polytomous models, Fisher information can be written as
\begin{equation}\label{eq:fisher_generic}
I_x(\theta)
=
\sum_{k=0}^{K-1}
\frac{\left(\frac{\partial}{\partial \theta} P(Y_x=k\mid\theta)\right)^2}{P(Y_x=k\mid\theta)}.
\end{equation}
MFI is computationally efficient but uses only a point estimate rather than the full posterior.

\subsubsection{Minimum Expected Posterior Variance (MEPV)}

MEPV is a Bayesian criterion that selects the item minimizing expected posterior variance after observing the (unknown) response:
\begin{equation}\label{eq:mepv}
x_{t+1}
=
\argmin_{x \in \mathcal I_{\mathrm{feas}}\setminus \mathcal I_t}
\mathbb E_{Y_x \mid \mathcal D_t}\!\left[\mathrm{Var}(\theta \mid \mathcal D_t, Y_x)\right].
\end{equation}
The expectation is computed under the posterior predictive distribution
\begin{equation}\label{eq:predictive}
P(Y_x=k \mid \mathcal D_t)
=
\sum_{g} w^{(g)}_t\, P(Y_x=k \mid \theta^{(g)}).
\end{equation}
For each possible response $k$, we form the hypothetical updated posterior via \eqref{eq:posterior_update}, compute its variance, and average over $k$. In our experiments, we use MEPV for 1D baselines.

\subsubsection{Multidimensional criteria}

For $D$-dimensional traits, Fisher information becomes a matrix $\mathbf I_x(\boldsymbol{\theta})$. We use an A-optimality-style Bayesian criterion that minimizes the expected trace of the posterior covariance matrix:
\[
x_{t+1}
=
\argmin_{x \in \mathcal I_{\mathrm{feas}}\setminus \mathcal I_t}
\mathbb E_{Y_x \mid \mathcal D_t}\!\left[\mathrm{tr}(\boldsymbol{\Sigma}_{t+1})\right],
\]
which reduces to MEPV when $D=1$.

\subsection{Prediction}

After $T$ adaptive queries, predictions for any target item $x$ use the posterior predictive distribution on the grid:
\begin{equation}\label{eq:cat_prediction}
P(Y_x=k \mid \mathcal D_T)
=
\sum_{g} w^{(g)}_T\, P(Y_x=k \mid \theta^{(g)}).
\end{equation}

\subsection{Implementation Details}\label{app:cat:implementation}

Table~\ref{tab:cat_hyperparams} summarizes hyperparameters used in our CAT implementations.

\begin{table}[ht]
\caption{Hyperparameters for CAT baseline implementations.}
\label{tab:cat_hyperparams}
\centering
\begin{tabular}{llc}
\toprule
\textbf{Parameter} & \textbf{Description} & \textbf{Value} \\
\midrule
\multicolumn{3}{l}{\textit{Grid-based posterior}} \\
$\theta_{\max}$ & Grid range: $\theta \in [-\theta_{\max}, \theta_{\max}]$ & 4.0 \\
$G$ (1D) & Number of grid points for GRM/GPCM & 41 \\
$G$ (MIRT) & Grid points per dimension for MGRM/MGPCM & 9 \\
$D$ & Latent dimensions for MIRT models & 3 \\
\midrule
\multicolumn{3}{l}{\textit{Parameter estimation (EM)}} \\
Max iterations & Maximum EM iterations & 50 \\
Tolerance & Convergence criterion (log-likelihood change) & $10^{-3}$ \\
\midrule
\multicolumn{3}{l}{\textit{Item selection}} \\
Criterion (1D) & Selection criterion for GRM/GPCM & MEPV \\
Criterion (MIRT) & Selection criterion for MGRM/MGPCM & A-optimality \\
\bottomrule
\end{tabular}
\end{table}

\end{document}